\documentclass[3p]{elsarticle}

\usepackage{hyperref}

\journal{Neurocomputing}

\usepackage{subcaption}	  
\usepackage{mwe}              
\usepackage{booktabs} 
\usepackage{bbm} 
\usepackage{comment} 
\usepackage{amsmath} 
\usepackage{amssymb} 
\usepackage{amsthm} 
\usepackage{xcolor} 

\newcommand{\E}[1]{\mathbb E \left[ #1 \right]}
\newcommand{\Var}[1]{\mathrm {Var} \left[ #1 \right]}
\newcommand{\Cov}[2]{\mathrm {Cov} \left[ #1, \; #2 \right]}
\newcommand{\Corr}[2]{\mathrm {Corr} \left[ #1, \; #2 \right]}
\newcommand{\Bias}[1]{\mathrm{Bias} \left[ #1 \right]}
\newcommand{\ECond}[2]{\mathbb E \left[ #1 \;\middle|\; #2 \right]}
\newcommand{\VarCond}[2]{\mathrm {Var} \left[ #1 \;\middle|\; #2 \right]}
\newcommand{\CovCond}[3]{\mathrm {Cov} \left[ #1, \; #2 \;\middle|\; #3 \right]}

\newcommand{\VarSmall}[1]{\mathrm {Var} [ #1 ]}

\newcommand{\Tr}[1]{\textnormal{Tr} \left [ #1 \right]}
\newcommand{\TrSmall}[1]{\textnormal{Tr}[ #1 ]}
\newcommand{\EstimVarf}{\hat \sigma^2_{\hat f}}
\newcommand{\EstimNoise}{\hat \sigma^2_{\varepsilon}}

\theoremstyle{plain}

\begin{document}

\begin{frontmatter}

\title{Uncertainty Quantification in Extreme Learning Machine: Analytical Developments, Variance Estimates and Confidence Intervals}


\author[mymainaddress]{Fabian Guignard\corref{mycorrespondingauthor}}
\cortext[mycorrespondingauthor]{Corresponding author}
\ead{fabian.guignard@unil.ch}

\author[mymainaddress]{Federico Amato}

\author[mymainaddress]{Mikhail Kanevski}

\address[mymainaddress]{Institute of Earth Surface Dynamics, Faculty of Geosciences and Environment,University of Lausanne, Switzerland}

\begin{abstract}
Uncertainty quantification is crucial to assess prediction quality of a machine learning model.
In the case of Extreme Learning Machines (ELM), most methods proposed in the literature make strong assumptions on the data, ignore the randomness of input weights or neglect the bias contribution in confidence interval estimations.
This paper presents novel estimations that overcome these constraints and improve the understanding of ELM variability. 
Analytical derivations are provided under general assumptions, supporting the identification and the interpretation of the contribution of different variability sources.
Under both homoskedasticity and heteroskedasticity, several variance estimates are proposed, investigated, and numerically tested, showing their effectiveness in replicating the expected variance behaviours.
Finally, the feasibility of confidence intervals estimation is discussed by adopting a critical approach, hence raising the awareness of ELM users concerning some of their pitfalls. 
The paper is accompanied with a scikit-learn compatible Python library enabling efficient computation of all estimates discussed herein. 
\end{abstract}


\end{frontmatter}


\section{Introduction}

Statistical accuracy measures such as variance, standard error and Confidence Intervals (CI) are crucial to assess the quality of a prediction. Model uncertainty quantification is needed to build CI and has a direct impact on the prediction interval, especially when dealing with small datasets \cite{Heskes}. 
Uncertainty quantities for Feed-forward Neural Networks (FNN) solving regression tasks can be obtained by means of different methods\cite{Tibshirani, dybowski2001confidence}.
Here these quantities are investigated in relationship with the use of the Extreme Learning Machine (ELM) model \cite{Huang2004}. ELM is a single-layer FNN with random input weights and biases, therefore allowing the optimization of the output weights through the Least Squares (LS) procedure.
One can think about ELM as a projection of inputs in a random feature space where a Multiple Linear Regression (MLR) with a null intercept is performed.

Three main uncertainty sources can be distinguished \cite{chatfield1995model}.
A first one comes from the data, and in particular from sampling variation and unexplained fluctuations, or noise. 
A second uncertainty source is related to the estimation of the model parameters, which in the case of an FNN correspond to the weights and biases \cite{dybowski2001confidence}. In ELM, input weights and biases are randomly chosen, which clearly generates uncertainty.
Moreover, despite being optimized through a procedure with a unique solution,
the estimation of the output weights depends on the random input weights and on the data, which therefore induces additional fluctuations. 
Finally, a third type of uncertainty source is due to the model structure. This source, generally referred at as structural uncertainty, is not considered in this paper.

A number of methods were proposed to obtain confidence or prediction intervals with ELM. 
A Bayesian formulation was introduced to integrate prior knowledge and produce directly CI \cite{Soria-Olivas2011, ning2015new}.
In the frequentist paradigm, bootstrap methods were investigated in the context of time series \cite{Wan2014}. 
Akusok et al.  proposed a method to estimate prediction intervals using a covariance matrix estimate coming from MLR \cite{Akusok2019}.

Most of these methods make Gaussian assumption on the output distribution or do not consider the bias in interval estimation, which may cause misleading conclusions. Moreover, resampling methods lead to important computational burden when the number of data is high.
Finally, it is often argued that randomness of the input weights and biases is supposed to be negligible providing the training set large enough. However, it is not always clear how many data is needed in practice.
Indeed, while stochastic input layer initialization can have weak impact in low dimension, it is still unclear what could happen when number of features or/and neurons are large. Because of the curse of dimensionality, the random drawing of the input weights and biases could have a higher impact than suspected. To further investigate such impact, the development of ELM variance estimation methods taking into account its stochastic nature is therefore extremely relevant.
Additionally, ELM is also used efficiently with small training dataset \citep{leuenberger2015elm} --- in which case precise variance estimate is crucial --- where the randomness of the input weights and biases should not be ignored. 

The contribution of this paper is threefold. First, analytical development are proposed to derive ELM variance taking into account also the contribution induced by the random input weights and biases.
This is done without any other assumptions on the noise distribution than the facts that it is centered and have a finite variance. In particular, the presented theoretical results hold for dependant and non-identically distributed data.
Second, homoskedastic and heteroskedastic variance estimates are provided, and some of their properties are investigated. While it may be argued that the homoskedastic case could be unrealistic, its study is of great interest as it provides an insightful propaedeutic value and develops the intuition for more advanced situations. Moreover, in case of applications with a small number of data, homoskedastic assumption may yield to better results.
Third, the paper proposes empirical bases to move towards CI estimations, which include the variability induced by the random input weights and biases. Their discussion will also raise the awareness of ELM users about some pitfalls of confidence interval estimations.
Overall, the results presented in this paper are expected to clarify the impact of input weights variability and noise, hence increasing the understanding of ELM variability. 

The remainder of the paper is organized as follows. Starting from general assumption on the noise covariance matrix, probabilistic formulas are derived for predicted output variance knowing the training input for single and ensemble of (regularized) ELM in section 2.
Based on these formulas, section 3 provides variance estimates when noise is independent with constant variance (homoskedastic case) and non-constant variance (heteroskedastic case), for which a Python implementation is available on GitHub, see the software availability at the end of the paper.
The effectiveness of the proposed estimates is demonstrated through numerical experiments in section 4, where estimation of CI is also discussed. Finally, section 5 concludes the paper.

\section{Analytical developments}

This section begins by recalling ELM theory and fixing notations. Subsequently, the bias and variance for a single ELM are derived. 
The results are then generalized to ELM ensemble. 
Finally, correlation between two ELMs is investigated.

\subsection{Background and notations}

Assume that an output variable $y$ depends of $d$ input variables $x_1, \dots, x_d$ through the relationship
\begin{equation*}\label{additive_regression}
y = f(\mathbf{x}) + \varepsilon (\mathbf{x}),
\end{equation*}
where $\mathbf{x} = (x_1, \dots, x_d)^T \in \mathbb{R}^d$ is the vector composed by the input variables, $f$ is a function of $\mathbf{x}$ whose value represent the deterministic part of $y$, and $\varepsilon (\mathbf{x})$ is a random noise depending on the input representing the stochastic part of $y$. 
It is assumed that, whatever the value of $\mathbf x$, the noise is 
centered and has a finite variance.

Let the training set $\mathcal D=\{ ( \mathbf x_i, y_i) : \mathbf x_i \in \mathbb R^d, y_i \in \mathbb R\}_{i=1}^n$ be a sample from the joint distribution of $(\mathbf{x}, y)$. 
Given a new input point $\mathbf x_0\in \mathbb R^d$, one wants to predict the corresponding output $y_0$. The value $f(\mathbf x_0)$ seems a good guess. However, the function $f$ is unknown.
The true function $f$ needs to be approximated based on the sample $\mathcal D$ in order to provide an estimate of the prediction $f(\mathbf{x}_0)$.

For convenience, the matrix composed by all training input points will be noted $X = [\mathbf x_1 \mid \dots \mid \mathbf x_n] \, \in \mathbb R^{d\times n}$.
Moreover, at the training points, the $n-$dimensional vectors
$\mathbf y = (y_1, \dots, y_n)^T$,
$\mathbf f=(f(\mathbf x_1), \dots, f(\mathbf x_n))^T$, and
$\boldsymbol \varepsilon = ( \varepsilon(\mathbf x_1), \dots,  \varepsilon(\mathbf x_n))^T$ are defined.  
The covariance matrix of $\boldsymbol \varepsilon$ knowing $X$ is noted $\Sigma \in \mathbb R^{n\times n}$. 

\subsubsection{Extreme Learning Machine}
\label{subsub:ELM}

ELM is a  single-layer FNN with a random initialization of the input weights $\mathbf w_j\in \mathbb R^d$ and biases $b_j\in \mathbb R$, for $j=1, \dots N$, where $N$ denotes the number of neurons of the hidden layer. All input weights and biases are independent and identically distributed (i.i.d.), and are generally sampled from a Gaussian or uniform distribution. They map the input space into a random feature space in a non-linear fashion
by-way-of the \emph{non-linear feature mapping} \citep{huang2015trends}
\[
\mathbf h(\mathbf x) =  
\begin{pmatrix}
h_1(\mathbf x), 
\dots, 
h_N(\mathbf x)\end{pmatrix}^T \in \mathbb R^N,
\]
where $h_j(\mathbf x) = g(\mathbf x ^T\mathbf w_j+ b_j)$ is the output of the $j$-th hidden node, for $j = 1, \dots, N$, and $g$ is any infinitely differentiable activation function \citep{huang2006extreme}. 
A $N-$hidden neurons ELM can generate output functions of the form
\[
f_N(\mathbf x) =  \mathbf h(\mathbf x)^T \boldsymbol\beta,
\]
where $\boldsymbol \beta \in \mathbb R^N$ is the vector of output weights that relate the hidden layer with the output node. 

The output weights are trained using the sample $\mathcal{D}$ and optimized regarding the $L_2$ criterion performing the following procedure.
The $n\times N$  hidden layer matrix, denoted $H$ and defined element-wise by $H_{ij} = h_j(\mathbf x_i)$, $i = 1, \dots, n,$ and $j = 1, \dots, N$, is computed.
Then the cost function
\[
E = ||\mathbf y - H\boldsymbol\beta||^2_2,
\]
is minimized, where $||\cdot||_2^2$ denotes the Euclidean norm. This is exactly the LS procedure for a design matrix $H$ \citep{davidson2004econometric, Davison}.
If the matrix $H^T H$ is of full rank and then invertible, the output weights are estimated as the classical linear regression, with the analytical solution of the minimization of $E$,
\[
\boldsymbol {\widehat \beta} = (H^T H)^{-1} H^T \mathbf y,
\]
where the matrix $(H^T H)^{-1} H^T$ is  the Moore-Penrose generalized inverse of the matrix $H$ and will be denoted $H^{\dagger}$ in the following.
Thus, ELM can be thought as a MLR with a null intercept, performed on regressors obtained by a random non-linear transformation of the input variables.

At a new point, the prediction is given by $\hat f (\mathbf x_0) = \mathbf h(\mathbf x_0)^T \boldsymbol {\widehat \beta}$. In the remainder of the paper, all dependencies in $\mathbf{x}_0$ will be dropped for convenience and the prediction will be noted  $\hat f (\mathbf x_0) = \mathbf h^T \boldsymbol {\widehat \beta}= \mathbf h^T H^\dagger \mathbf y$.
The vector of model predictions at training points will be noted
$\mathbf {\hat{\text{$\mathbf f$}}}$, defined element-wise by $\mathbf{\hat{\text{$\mathbf f$}}}_i = \hat f(\mathbf x_i)$. 
The (random) matrix of all input weights and biases will be denoted 
\[
W = 
\left[
\begin{array}{c|c|c}
\mathbf w_1 & \dots & \mathbf w_N \\
\hline
b_1, & \dots & b_N, 
\end{array}
\right].
\]

\subsubsection{Regularized Extreme Learning Machine} 

To avoid overfitting and reduce outlier effects, a regularized version of ELM was proposed \cite{deng2009regularized}.
Highly variable output weights due to multicolinearity among neurons can be stabilized with regularization, too.
As mentioned by \cite{lendasse2013extreme}, this model is basically a --- potentially weighted --- \emph{Tikhonov regularization}, also known as \emph{ridge regression} \citep{piegorsch2015statistical}.
The output weights are optimized regarding the cost function
\[
E = ||\mathbf y - H\boldsymbol {\beta}||^2_2 + \alpha||\boldsymbol {\boldsymbol {\beta}}||^2_2,
\]
for some real number $\alpha >0$, sometimes called the \emph{Tikhonov factor}, which controls the penalization of taking big output weights. Noting $I$ the identity matrix, the analytical solution of this optimization problem  for a fixed $\alpha$ is given by
\[
\boldsymbol {\widehat \beta} = (H^T H + \alpha I)^{-1} H^T \mathbf{y},
\]
thanks to the fact that the matrix $(H^T H + \alpha I)$ is always invertible, see \cite{boyd2018introduction}.
To lighten the notation, the matrix $H^\alpha = (H^T H + \alpha I)^{-1} H^T$ is defined.
Remark that as $\alpha$ goes to zero, $H^{\alpha}$ goes to $H^{\dagger}$ and the classical ELM is recovered \cite{deng2009regularized}.
In the remainder of the paper, most of results are presented with $H^\alpha$, but they remain valid for the non-regularized case, unless the contrary is clearly specified. 

\subsubsection{Extreme Learning Machine Ensemble} 
Another way to avoid overfitting is to combine several ELM models. This also reduce the randomness induced by the input weight initialization, which could be beneficial --- especially for small datasets. 
Several ensemble techniques have been developed for ELM
\citep{liu2010ensemble, lendasse2013extreme}.
In this paper, each model of a given ensemble will have the same activation function and number of neurons, and all models will be averaged after training.
This corresponds to retrain $M$ times the model and average the results, where $M$ is the number of ELM networks in the ensemble.
The hidden layer matrix and the matrix of input weights and biases of the $m-$th retraining will be noted respectively $H_m$ and $W_m$, for $m = 1, \dots, M$. If the $m-$th prediction is noted $\hat f_m(\mathbf x_0)$, the final prediction is
\begin{equation*}
\label{equ:relm}
\hat f(\mathbf x_0) 
= \frac{1}{M}\sum_{m=1}^M \hat f_m(\mathbf x_0) 
= \frac{1}{M}\sum_{m=1}^M \mathbf h_m^T H_m^{\alpha}\mathbf y,
\end{equation*}
where $\mathbf h_m$ and $H_m^{\alpha}$ are the analogous quantities defined previously for the $m$-th model. Note that the weights have the same joint distribution across all the models, which allows us to drop the $m$ index in most calculations of the remainder of this paper. 

\subsection{Bias and variance for a single ELM}

In this section, the uncertainty for (regularized) ELM is explored. For all derivations, it is supposed that hyper-parameters $N$ and $\alpha$ --- when applicable --- are considered as fixed and non-stochastic. Also, all formulas are derived knowing $X$. However, this conditioning is dropped to avoid cumbersome notations.
Note that  $\hat f (\mathbf x_0)$ is a random variable depending on the noise at training points $\boldsymbol \varepsilon$, but also on the input weights and biases $W$ used in the construction of $\mathbf h$ and $H^\alpha$.
As the noise is centred, 
\begin{equation}\label{equ:linbias}
\ECond{\hat f (\mathbf x_0) }{W} = 
\mathbf h ^T  H^{\alpha} \E{\mathbf f + \boldsymbol \varepsilon} = \mathbf h ^T H^{\alpha} \mathbf f.
\end{equation}
Using the law of total expectation, one can compute the bias of the model at $\mathbf x_0$,
\begin{equation*}
    \label{equ:bias1}
    \Bias{\hat f (\mathbf x_0)} = 
    \E {\ECond{\hat f (\mathbf x_0) }{ W}} - f (\mathbf x_0) 
    = \E{\mathbf h ^T H^{\alpha} \mathbf f } - f (\mathbf x_0).
\end{equation*}

Let us now compute the variance of the model at a new point. First, one have
\begin{equation}\label{equ:linvar}
\VarCond{\hat f(\mathbf x_0)}{ W } = 
\mathbf h^TH^{\alpha}  \Var{\mathbf f + \boldsymbol \varepsilon} H^{\alpha T} \mathbf h   = \mathbf h^TH^{\alpha}  \Sigma H^{\alpha T} \mathbf h, 
\end{equation}
which is the typical variance expression for MLR.
With equations \eqref{equ:linbias} and \eqref{equ:linvar}, the variance of the model at $\mathbf x_0$ can be computed by using the law of total variance,
\begin{align}
\label{equ:modvar1}
\begin{split}
\Var{\hat f (\mathbf x_0) }
&= \E{\VarCond{\hat f (\mathbf x_0)}{W}} + \Var{ \ECond{\hat f (\mathbf x_0)}{ W} }\\ 
&= \E { \mathbf h^TH^{\alpha}  \Sigma H^{\alpha T} \mathbf h } 
+ \Var {\mathbf h^T H^{\alpha} \mathbf f}.
\end{split}
\end{align}
The first term of the right-hand side (RHS) is the variance of the LS step averaged on all possible random feature spaces generated by input weights and biases, while the second term is the variation of the LS step bias across all random feature spaces. Note that the second term appears if and only if the random input weights and biases are considered. 
In the non-regularized case with independent homoskedastic noise, if $W$ and $X$ are deterministic, the classical MLR formula for the variance at a prediction point is recovered, see \cite{Davison}. 

\subsection{Bias and variance for ELM ensemble}

As mentioned before, the training could be done several times and averaged.
A direct calculation --- which can be found in the appendix --- can be done for bias and variance of the averaged predictor. 
Basically, it uses the law of total variance and elementary probability calculus from which one get
\begin{equation}
    \label{varELME}
    \Var{\hat f (\mathbf x_0)}
    =   \frac{1}{M}  \E{ \mathbf h^TH^{\alpha}  \Sigma H^{\alpha T} \mathbf h }
    +\frac{M- 1}{M}  \E{ \mathbf h^T H^{\alpha}} \Sigma \,
    \E{ H^{\alpha T} \mathbf h }
    +\frac{1}{M} \Var{\mathbf h^T H^{\alpha} \mathbf f },
\end{equation}
while the bias still unchanged.
The RHS first and third terms are the single ELM variance divided by the number of models. 
The bias variation of the LS step is reduced by a $1/M$ factor. 
Although the average variance of the LS step represented by the RHS first term seems to decrease by a $1/M$ factor, models are pairwise dependent which yields the RHS second term.
Notice that if $M=1$, equation \eqref{equ:modvar1} is recovered. 
If $M$ grows, the RHS second term tends to dominate the model variance.
Remark also that using the law of total covariance, it is easily checked by analogous computation that the covariance between two members of an ELM ensemble correspond to $\E{ \mathbf h^T H^{\alpha}} \Sigma \,
\E{ H^{\alpha T} \mathbf h}$.

\subsection{Use of random variable quadratic forms}

Formulation of variance in equations \eqref{equ:modvar1} and \eqref{varELME} are convenient for the interpretation of ELM as a MLR on random features. However, quadratic forms in random variables appears in these formulas, which allows to pursue calculations. With the Corollary 3.2b.1 of \cite{mathai1992quadratic}, the expectation of random variable quadratic form can be computed as the quadratic form in its expected values plus the trace of its covariance matrix times the matrix of the quadratic form. This Corollary will be used extensively in this paper, each time an expectation of a quadratic form in random variables appears.

Setting $\mathbf z = H^{\alpha T} \mathbf h$ and
assuming the existence of its expectation $\boldsymbol{\mu}$ and covariance matrix  $C$, equation \eqref{equ:modvar1} becomes
\begin{align*}
\begin{split}
    \Var{\hat f (\mathbf x_0)}
    &=\Tr{\Sigma C} + \boldsymbol{\mu}^T\Sigma \boldsymbol{\mu} + \mathbf f^T  C \mathbf f,
\end{split}
\end{align*}
where $\Tr{\cdot}$ denotes the trace of a square matrix.  Although the notation do not specify it, the quantities $\mathbf z, \boldsymbol{\mu}$ and $C$ depend on the Tikhonov factor $\alpha$ in the regularized case. Similarly, $\mathbf z_m = H_m^{\alpha T} \mathbf{h}_m$ is set for ELM ensembles and the  variance becomes
\begin{equation} \label{equ:modvarensemblequad}
    \Var{\hat f (\mathbf x_0)}
    = \frac{1}{M}\Tr{\Sigma C} + \boldsymbol{\mu}^T \Sigma \boldsymbol{\mu}+ \frac{1}{M} \mathbf f^T  C \mathbf f. 
\end{equation}

\subsection{Correlation between two ELMs}

As the covariance between two single ELMs is $\boldsymbol{\mu}^T\Sigma \boldsymbol{\mu}$, their linear correlation at $\mathbf{x}_0$  is given by
\begin{align*}
\begin{split}
    \Corr{\hat f_1 (\mathbf x_0)} {\hat f_2 (\mathbf x_0)}
    &=\frac{\boldsymbol{\mu}^T\Sigma \boldsymbol{\mu}}{\Tr{\Sigma C} + \boldsymbol{\mu}^T\Sigma \boldsymbol{\mu} + \mathbf f^T  C \mathbf f}.
\end{split}
\end{align*}
Remark that considering the input weights and biases as fixed is equivalent to ignore $\Tr{\Sigma C} + \mathbf f^T  C \mathbf f$ and to have a correlation of 1 between the two models. 

An interesting insight is provided by the case of independent and homoskedastic noise, i.e. $\Sigma = \sigma^2_{\varepsilon}\, I$, which yields
\begin{align*}
\begin{split}
    \Corr{\hat f_1 (\mathbf x_0)} {\hat f_2 (\mathbf x_0)}
    &=\frac{\boldsymbol{\mu}^T \boldsymbol{\mu}}{\Tr{C} +  \boldsymbol{\mu}^T\boldsymbol{\mu} + \frac{1}{\sigma^2_{\varepsilon}} \mathbf f^T  C \mathbf f}.
\end{split}
\end{align*}
Notice that in this particular case, when $\sigma^2_{\varepsilon}$ is small the linear correlation between two ELMs vanishes. Contrariwise, when  $\sigma^2_{\varepsilon}$ is large the linear correlation between two ELMs tends to $b = \boldsymbol{\mu}^T \boldsymbol{\mu} / (\boldsymbol{\mu}^T \boldsymbol{\mu} + \Tr{C})$. Therefore, the amount of noise has a direct impact on the linear correlation which takes its value between $0$ and $b\leq 1$.
The trace of $C$ can be interpreted as a variability measure of $\mathbf{z}$, called sometimes the \textit{total variation} or the \textit{total dispersion} of $\mathbf{z}$ \citep{seber2009multivariate}. In our case, it controls the linear correlation bound $b$, in the sense that more variable is $\mathbf z$, farther from 1 is the maximal value that the linear correlation can take, regardless the noise in the data.

\section{Variance estimation of ELM ensemble}
\label{sec:estimate}

This section introduces novel estimates of the ELM variance. Although several ELM are necessary to allow the estimation of quantities related to $\mathbf z$ --- which motivates the use of ELM ensembles --- reliable results are also obtained with very small ensemble. First, the variation of the LS step bias overall random feature space is estimated. Then, assuming noise independence, the variance of the LS step averaged on all possible random feature spaces is estimated under homoskedasticity and heteroskedasticity for non-regularized and regularized ELM ensembles.

\subsection{Estimation of the least squares bias variation}

The quantity $\mathbf f^T  C \mathbf f = \Var{\mathbf h^T H^{\alpha} \mathbf f} $ doesn't depends on noise. It is the variance induced by the randomness of $W$, knowing the true function $f$ at training points. Tentatively assume that the output weights are not regularized. As $\mathbf{f}$ is unknown, one  approximate it by the model prediction at the training points $\mathbf {\hat{\text{$\mathbf f$}}} = H H^{\dagger} \mathbf y$. For each model of the ensemble, 
\begin{equation*} 
\Var{\mathbf h^T H^{\dagger} \mathbf f}  
\approx \VarCond{\mathbf h^T H^{\dagger} \mathbf{\hat{\text{$\mathbf f$}}}}{ \boldsymbol \varepsilon}
 =\VarCond {\hat f (\mathbf x_0 )}{  \boldsymbol \varepsilon}.
\end{equation*}
This motivate the following estimate for $\mathbf f^T  C\mathbf f$,
\begin{equation}\label{equ:estimLSbiasvariation}
     \EstimVarf =  \frac{1}{M-1}\sum_{m=1}^M \left ( \hat f_m(\mathbf x_0) - \frac{1}{M}\sum_{l=1}^M \hat f_l(\mathbf x_0) \right)^2.
\end{equation}
The same estimate will be used for the regularized case.

The expectancy of $\EstimVarf$ can be easily computed. Knowing the noise at the training points,
\begin{align*}
\begin{split}
\ECond{\EstimVarf}{\boldsymbol{\varepsilon}} 
&= \VarCond{\hat f(\mathbf{x}_0)}{\boldsymbol{\varepsilon}} = \VarCond{\mathbf{z}^T\mathbf{y}}{\boldsymbol{\varepsilon}} \\
&= \VarCond{\mathbf{z}^T \mathbf{f}}{\boldsymbol{\varepsilon}} + \VarCond{\mathbf{z}^T \boldsymbol{\varepsilon}}{\boldsymbol{\varepsilon}} +
2 \CovCond{\mathbf{z}^T \mathbf{f}}{\mathbf{z}^T \boldsymbol{\varepsilon}}{\boldsymbol{\varepsilon}} \\
&= \Var{\mathbf{z}^T \mathbf{f}} + \boldsymbol{\varepsilon}^T C \boldsymbol{\varepsilon} +
2 \mathbf{f}^T C \boldsymbol{\varepsilon},
\end{split}
\end{align*}
using the unbiasedness of the estimate in the first equality. 
By taking the expectation over $\boldsymbol{\varepsilon}$ on both side, 
\begin{align*}
\begin{split}
\E{\EstimVarf} = \Var{\mathbf h^T H^{\alpha}\mathbf{f}} + \Tr{ \Sigma C}.
\end{split}
\end{align*}
This shows that the bias of the estimate defined in equation \eqref{equ:estimLSbiasvariation} is given by $\Tr{ \Sigma C }$, which is the first term of the RHS of equation \eqref{equ:modvarensemblequad} up to a factor $1/M$. Therefore, regardless of a particular form of $\Sigma$ or whether ELM is regularized or not,  it is unnecessary to estimate the latter and $(1/M)\, \EstimVarf$ is an unbiased estimate of the sum of the first and last terms of equation \eqref{equ:modvarensemblequad}.

\subsection{Estimation under independence and homoskedastic assumptions}
\label{subsec:homo}

Only the second term of RHS of equation \eqref{equ:modvarensemblequad} remains to be estimated.
If the noise is assumed to be independent and have a constant variance,  the covariance matrix of $\boldsymbol{\varepsilon}$ writes $\Sigma = \sigma^2_{\varepsilon}\, I$ and the second term of RHS of equation \eqref{equ:modvarensemblequad} becomes 
$ \boldsymbol{\mu}^T \Sigma \boldsymbol{\mu} = \sigma^2_{\varepsilon} \, \boldsymbol{\mu}^T  \boldsymbol{\mu}$. The quantity $\boldsymbol{\mu}^T \boldsymbol{\mu}$ --- which, knowing $X$, stochastically depends only on $W$ --- will be estimated separately from $\sigma_\varepsilon^2$.

\subsubsection{Estimation of $\boldsymbol{\mu}^T \boldsymbol{\mu}$}
\label{subsubsec: muTmu}

As a first step,   $\boldsymbol{\mu}^T \boldsymbol{\mu}$ is naively estimated with 
\begin{equation} \label{estim:muTmu}
  \widehat{\boldsymbol{\mu}}^T  \widehat{\boldsymbol{\mu}} = \frac{1}{M^2} \sum_{m, l=1}^M \mathbf z^T_m \mathbf z_l,
 \quad
 \text{ where }
 \quad
 \widehat{\boldsymbol{\mu}} = \frac{1}{M} \sum_{m=1}^M \mathbf z_m.   
\end{equation}
However, remark that
\begin{align*}
\begin{split}
 M^2 \E{
 \widehat{\boldsymbol{\mu}}^T  \widehat{\boldsymbol{\mu}}
 }
 & =
\sum_{m=1}^M 
  \E{
  \mathbf z^T_m
\mathbf z_m
} +
\sum_{m \neq l} 
  \E{\mathbf z_m}^T
  \E{\mathbf z_l}\\
  &= M \left(\Tr{C} + \boldsymbol{\mu}^T \boldsymbol{\mu} \right) + M(M-1)\boldsymbol{\mu}^T \boldsymbol{\mu} \\
  &=
  M\, \Tr{C} + M^2 \,\boldsymbol{\mu}^T \boldsymbol{\mu},
\end{split}
\end{align*}
and dividing by $M^2$ shows that the estimate given in \eqref{estim:muTmu} has a bias equal to $(1/M)\Tr{C}$, which comes from the expected values of the $M$ quadratic terms in $\mathbf{z}_m$.

To remove this bias, one can estimate it by
\begin{equation} \label{estim:traceC}
\frac{1}{M}\TrSmall{\widehat{Q}},
\quad
\text{ where }
\quad
\widehat{Q}  = \frac{1}{M-1} \sum_{m=1}^{M} \left(\mathbf{z}_m- \widehat{\boldsymbol{\mu}}\right) \left(\mathbf{z}_m- \widehat{\boldsymbol{\mu}}\right)^T.  
\end{equation}
This is an unbiased estimate of $(1/M)\Tr{C}$, which immediately follows from the fact that $\widehat{Q}$ is an unbiased estimate of $C$. Therefore, subtract \eqref{estim:traceC} from \eqref{estim:muTmu} yields an unbiased estimate of $\boldsymbol{\mu}^T  \boldsymbol{\mu}$.
Note that
\begin{align*}
\begin{split}
    (M-1)\,\TrSmall{\widehat{Q}} 
    &=
    \sum_{m=1}^{M} \left(\mathbf{z}_m- \widehat{\boldsymbol{\mu}}\right)^T \left(\mathbf{z}_m- \widehat{\boldsymbol{\mu}} \right)\\
    &=
    \sum_{m=1}^{M} \mathbf{z}_m^T \mathbf{z}_m 
    - M \, \widehat{\boldsymbol{\mu}}^T \widehat{\boldsymbol{\mu}}. \\\end{split}
\end{align*}
Hence, the unbiased estimate of $\boldsymbol{\mu}^T  \boldsymbol{\mu}$ that was just developed results in
\begin{align}\label{estim:algohomo}
\begin{split}
    \widehat{\boldsymbol{\mu}}^T  \widehat{\boldsymbol{\mu}} - \frac{1}{M}\TrSmall{\widehat{Q}} 
    &= 
    \widehat{\boldsymbol{\mu}}^T  \widehat{\boldsymbol{\mu}}
    - \frac{1}{M(M-1)} \sum_{m=1}^{M} \mathbf z^T_m
    \mathbf z_m 
    +\frac{1}{M-1}    \widehat{\boldsymbol{\mu}}^T  \widehat{\boldsymbol{\mu}} \\
    &=  
    \frac{M}{M-1}
    \widehat{\boldsymbol{\mu}}^T  \widehat{\boldsymbol{\mu}}
    -  \frac{1}{M(M-1)} \sum_{m=1}^{M} \mathbf z^T_m
    \mathbf z_m. 
\end{split}    
\end{align}
Substituting equation \eqref{estim:muTmu} into equation \eqref{estim:algohomo}, the computation still goes on, and
\begin{equation} \label{eq:homo_diagoff}
    \widehat{\boldsymbol{\mu}}^T \widehat{\boldsymbol{\mu}} - \frac{1}{M}\TrSmall{\widehat{Q}} 
    = \frac{1}{M(M-1)}\sum_{m \neq l} 
  \mathbf z^T_m
    \mathbf z_l. 
\end{equation}
This shows that the estimate \eqref{eq:homo_diagoff} removes the quadratic terms $m=l$ from which 
the bias of the naive estimate \eqref{estim:muTmu} was induced.
However, note that formulation \eqref{estim:algohomo} is more convenient to compute than \eqref{eq:homo_diagoff}, from an algorithmic perspective.

\subsubsection{Noise estimation}

The estimation of $\sigma_\varepsilon^2$ is separated in two cases, the non-regularized and the regularized ones.
A couple of notations is needed to make readable the equations. The residuals for the $m$-th model are $\mathbf{r}_m= \mathbf{y} - P_m\mathbf{y}$, where $P_m = H_m H_m^{\alpha}$. 
Also, $\mathbf{b} =  \mathbf{f} - \mathbb{E}[\mathbf{\hat f}]  = ( I - \E{P} )\mathbf {f}$ is the vector of bias of the model predictions at the training points $\mathbf{\hat f}$, 
and
$\mathbf{b}_{w,m} = 
\mathbf{f} - \mathbb{E}[\mathbf{\hat f}\;|\;W_m] = (I - P_m) \mathbf{f}$ is the vector of conditional bias of the model predictions at the training points $\mathbf{\hat f}$ for the $m$-th model knowing $W_m$.

Let us first concentrate on the non-regularized case. Then, $P_m$ is a projection matrix.
A natural way to obtain estimate for $\EstimNoise$ is to start with the expectation of the residual sum of squares (RSS) based on the averaged ensemble. However, mainly due to the fact that the expectation of a projection matrix is not a projection matrix, it is preferred here to work with the RSS of each model.
Using this, the expectation of the residual sum of squares for the $m$-th ELM knowing input weights and biases is given by
\begin{align}\label{eq:ERSSW_homo}
\begin{split}
    \ECond{\mathbf{r}_m^T
    \mathbf{r}_m}{W_m} 
    &= \ECond{\mathbf{y}^T (I -  P_m)\mathbf{y}}{W_m} \\
    &= \sigma^2_\varepsilon  \,\Tr{ I -  P_m} + \mathbf{f}^T (I - P_m) \mathbf{f} \\
    &= \sigma^2_\varepsilon (n-N) + \mathbf{f}^T \mathbf{b}_{w,m},
\end{split}
\end{align}
and taking the expectation over the input weights and biases yields
$\mathbb{E}[\mathbf{r}_m^T\mathbf{r}_m]
= \sigma^2_\varepsilon \left(n - N\right) + \mathbf{f}^T \mathbf{b}$.
This motivates the following estimate, 
\begin{equation}\label{eq:EstimNoise}
    \EstimNoise = \frac{1}{M(n-N)}\sum_{m=1}^M \mathbf{r}_m^T\mathbf{r}_m,
\end{equation}
which is the average of all MLR estimates of $\sigma^2_\varepsilon$. Its bias is directly obtained from previous calculation, yielding
\begin{equation} \label{eq:HomoNoiseBias}
    \Bias{\EstimNoise} = \frac{1}{n - N}\, \mathbf{f}^T \mathbf{b} \; \geq 0 \; ,
\end{equation}
where non-negativity results from the facts that a projection matrix is positive semidefinite and expectation of a positive semidefinite matrix is positive semidefinite.

If regularized ELMs are used, $P_m$ is no more a projection matrix,
and the expected RSS for each ELM of the ensemble knowing $W_m$ becomes
\begin{align} \label{ERSSW_homo_reg}
\begin{split}
    \ECond{\mathbf{r}_m^T \mathbf{r}_m}{W_m} 
    &= \ECond{\mathbf{y}^T (I - P_m)^T(I -  P_m)\mathbf{y}}{W_m} \\
    &= \sigma^2_\varepsilon\, \Tr{(I - P_m)^T(I -  P_m)} + \mathbf{f}^T (I - P_m)^T(I - P_m) \mathbf{f}\\
    &= \sigma^2_\varepsilon \, \left(n - 2 \Tr{P_m} + \Tr{P_m^2}\right) + \mathbf{b}_{w,m}^{T}\mathbf{b}_{w,m}.
\end{split}
\end{align}
Analogously to what is done in \cite{hastie1990generalized}, the \emph{effective degrees of freedom for error} can be defined as $n-\gamma$, with $\gamma = 2 \E{\Tr{ P}} - \E{\Tr{P^2}}$.
Expectation over input weights and biases of equation \eqref{ERSSW_homo_reg} gives
$
\E{\mathbf{r}_m^T\mathbf{r}_m} 
= \sigma^2_\varepsilon \left(n - \gamma\right) + \mathbb{E}[\mathbf{b}_{w}^{T}\mathbf{b}_{w}]
$
, hence,
\begin{align*}
\begin{split}
    \E{\frac{1}{M(n-\gamma)}\sum_{m=1}^M\mathbf{r}_m^T\mathbf{r}_m} 
    &= \sigma^2_\varepsilon  + \frac{1}{n-\gamma}\E{ \mathbf{b}_{w}^{T}\mathbf{b}_{w}}\\
    &= \sigma^2_\varepsilon  + \frac{1}{n-\gamma}\left(\mathbf{b}^T \mathbf{b} + \Tr{\Var{ \mathbf{b}_{w}}}\right),
\end{split}
\end{align*}
which make appears the squared bias and the total variation of the conditional bias.
This motivates the following estimate in the regularized case,
\begin{equation}\label{eq:EstimNoiseReg}
    \EstimNoise = \frac{1}{M(n-\widehat{\gamma})}\sum_{m=1}^M \mathbf{r}_m^T\mathbf{r}_m,
    \quad\text{ with }\quad
    \widehat \gamma = \frac{1}{M} \sum_{m=1}^M \left( 
    2 \Tr{P_m} - \Tr{P_m^2} \right),
\end{equation}
and it is easy to check that
\begin{align}\label{eq:biasreg}
\begin{split}
    \Bias{\EstimNoise}
    &=\E{\frac{1}{n-\widehat \gamma }\,\mathbf{b}_{w}^{T}\mathbf{b}_{w}}\; \geq 0 \; .
\end{split}
\end{align}

Computationally,  $\hat \gamma$ can be efficiently calculated using the singular value decomposition of $H_m$. Indeed, it can be shown \citep{piegorsch2015statistical, golub1979generalized} that the trace of $P_m$ and $P_m^2$ are given by
\begin{equation}\label{eq:TraceComputation}
\Tr{P_m} = \sum_{i=1}^N \frac{\lambda_{m,i}}{\lambda_{m,i} + \alpha} 
\quad
\text{ and }
\quad
\Tr{P_m^2} = \sum_{i=1}^N \left(\frac{\lambda_{m,i}}{\lambda_{m,i} + \alpha} \right)^2,
\end{equation}
where $\lambda_{m, i}$, $i = 1, \dots, N$ are the eigenvalues  of $H_m^T H_m$.
In particular, substitution of  \eqref{eq:TraceComputation} in \eqref{eq:EstimNoiseReg} and elementary manipulations allows to writes  
\[
\hat \gamma = N - \frac{1}{M}\sum_{m=1}^M\sum_{i=1}^N \left(\frac{\alpha}{\lambda_{m,i} + \alpha} \right)^2
\]
where $\sqrt{\lambda_{m,i}}$,  $i = 1, \dots, N$ are the singular values of $H_m$. Note that this latter equation also show the drop in the degrees of freedom lost due to regularization, comparing to the non-regularized case.

\subsubsection{Estimations of ELM ensemble variance}

In \cite{guignard2020model}, the authors proposed --- only for the non-regularized case --- the following naive homoskedastic estimate of the variance of $\hat f(\mathbf{x}_0)$,
\[
\hat \sigma^2_{NHo} =  \EstimNoise \,  \widehat{\boldsymbol{\mu}}^T  \widehat{\boldsymbol{\mu}}
+ \frac{1}{M}\EstimVarf.
\]
This naive estimate directly use equation \eqref{estim:muTmu} to approximate $\boldsymbol{\mu}^T  \boldsymbol{\mu}$ without considering its bias. However, a bias-reduced estimate is obtained by estimating $\boldsymbol{\mu}^T  \boldsymbol{\mu}$ by equation \eqref{eq:homo_diagoff}, yielding
\[
\hat \sigma^2_{BR} =  \EstimNoise \,  \left(\widehat{\boldsymbol{\mu}}^T  \widehat{\boldsymbol{\mu}} -\frac{1}{M}\TrSmall{\hat Q}    \right)
+ \frac{1}{M}\EstimVarf.
\]
Using the covariance definition, it is easy to see that
\begin{equation}\label{eq:final bias}
\Bias{\hat \sigma^2_{BR}} = 
\Bias{\EstimNoise }  
\boldsymbol{\mu}^T \boldsymbol{\mu}
+ \Cov{\EstimNoise}{\widehat{\boldsymbol{\mu}}^T  \widehat{\boldsymbol{\mu}} -\frac{1}{M}\TrSmall{\hat Q}}, 
\end{equation}
where the bias of $\EstimNoise$ is \eqref{eq:HomoNoiseBias} or \eqref{eq:biasreg}.
Note that in both cases, while the first term of the RHS of equation \eqref{eq:final bias} is always non-negative, the estimates of $\sigma_\varepsilon^2$ and $ \boldsymbol{\mu}^{T}\boldsymbol{\mu}$ could be correlated, introducing the second term. However, this supplementary bias could be negative, potentially compensating the first term. Note that its magnitude is bounded by 
\begin{equation}\label{eq:CovBound}
    \left| \;
    \Cov{
    \EstimNoise }{\widehat{\boldsymbol{\mu}}^T  \widehat{\boldsymbol{\mu}} -\frac{1}{M}\TrSmall{\hat Q} }\; \right|
    \leq
    \frac{2}{M(n-N)}
    \sqrt{\mathbf{f}^T\Var{\mathbf{b}_{w}} \mathbf{f}  \, \cdot \,
    \boldsymbol{\mu}^T C \boldsymbol{\mu}},
\end{equation}
in the non-regularized case --- see the appendix --- showing that this covariance term vanishes with large $M$.
Remark also that the bias of  $\hat \sigma^2_{NHo}$ has an additional non-negative term, $ (1/M)\Tr{C}\E{\EstimNoise }$, which disappears in \eqref{eq:final bias} thanks to the unbiasedness of $\widehat{\boldsymbol{\mu}}^T  \widehat{\boldsymbol{\mu}} -(1/M)\TrSmall{\hat Q} $.  

$\quad$ 

\subsection{Estimation under independence and heteroskedastic assumptions}
\label{subsec:hetero}

Suppose the noise is  independent but have variance which have a dependence of unknown form  on $\mathbf x$.
Then $ S=\boldsymbol{\mu}^T \Sigma \boldsymbol{\mu}$ has to be estimated considering the noise covariance matrix $\Sigma$ as diagonal.
To this aim, it could be possible to reuse estimates from MLR. 
However, several estimates are based on the evaluation of the covariance matrix $ H_m^{\alpha }  \Sigma_m  H_m^{\alpha T}$ of the output weights $\boldsymbol{\widehat{ \beta}}_m$. In this paper, the modified heteroskedastic-consistent covariance matrix estimator (HCCME) obtained from the (ordinary) Jackknife \citep{mackinnon1985hc} --- noted $\text{HC}_3$ and extended to the ridge regression case \citep{nyquist1988ridgejack} --- is used,
\begin{equation} \label{eq:HCCME}
    \text{HC}_3 = H_m^{\alpha}\widehat{\Sigma}_m H_m^{\alpha T},
    \quad \text{ with } \quad \widehat{\Sigma}_m = \frac{n-1}{n} \left[ \widehat{\Omega}_m - \frac{1}{n}\tilde{\mathbf r}_m\tilde{\mathbf r}_m^T \right] ,
\end{equation}
where $\tilde{\mathbf r}_m$ is the vector defined element-wise by $\tilde{\mathbf r}_{m, i} = \mathbf r_{m, i}/(1-p_{m,i})$, $p_{m,i}$ is the $i$-th diagonal element of $P_m$ and 
$\widehat{\Omega}_m$ is the diagonal matrix with the $i$-th diagonal element equal to $\tilde{\mathbf r}^2_{m, i}$.
This estimate is still valid for the non-regularized case, for which --- under some technical assumptions --- it is consistent \citep{white1980heteroskedasticity, mackinnon1985hc}, while $\hat \Sigma_m$ is an inconsistent estimator of $\Sigma$.
Nevertheless, other HCCME estimates could be used, such as $\text{HC}_0$ \citep{white1980heteroskedasticity}, $\text{HC}_1$ got from the weighted Jackknife \citep{hinkley1977jackknifing}, $\text{HC}_2$ proposed in \cite{horn1975estimating} or $\text{HC}_4$ proposed more recently in \cite{cribari2004asymptotic}. The HC notation follows what can be found in \cite{davidson2004econometric}, which provides useful insight on this kind of estimators.
Note that for sufficiently large $n$, $\text{HC}_3$ is close to $H_m^{\alpha}\widehat{\Omega}_m H_m^{\alpha T}$, which corresponds to the estimate used in \cite{Akusok2019} to build prediction intervals for large amounts of data, assuming fixed input weights.

If a unique ELM model is performed, the use of the HCCME is straightforward. Nonetheless, as one attempts to take into account the input weight variability through ELM replications, the HCCME is applied in three different ways.
Suppose first that $\Sigma$ is known and write the estimate
 \begin{equation}\label{eq:invest_intuition}
    \widehat{\boldsymbol{\mu}}^T \Sigma  \widehat{\boldsymbol{\mu}}
 = \frac{1}{M^2} \sum_{m, l=1}^M \mathbf z^T_m \Sigma \mathbf z_l
  = \frac{1}{M^2} \sum_{m, l=1}^M  \mathbf h_m^T H_m^{\alpha }  \Sigma H_l^{\alpha T} \mathbf h_l.
 \end{equation}
Inspecting equations \eqref{eq:HCCME} and \eqref{eq:invest_intuition}, a first natural suggestion is to estimate $ \boldsymbol{\mu}^T \Sigma \boldsymbol{\mu}$ with
\[
\widehat S_1 = \frac{1}{M}\sum_{m=1}^M \mathbf z^T_m  \widehat{\Sigma}_m \mathbf z_m.
\]
Note that $\widehat S_1$ estimates the covariance matrix of the output weights with the HCCME for each of the $M$ random feature spaces. Although it has the advantage of reusing the HCCME in its original formulation, the quadratic forms in random vectors depending on input weight may produce an additional bias.
Another estimator is obtained by naively evaluating all cross terms of equation \eqref{eq:invest_intuition},
\[
\widehat S_{NHe} = \frac{1}{M^2}\sum_{m, l=1}^M \mathbf z^T_m \widehat{\Sigma}_m \mathbf z_l.
\]
Note that this is equivalent to estimate $\Sigma$ with $(\widehat\Sigma_{m}+\widehat\Sigma_{l})/2$ in equation \eqref{eq:invest_intuition}, see \cite{guignard2020model}.
However, terms for which $m=l$ may produce additional biases, similarly to what was shown for the homoskedastic estimate of equation  \eqref{estim:muTmu}, which motivates
\[
\widehat S_2 = \frac{1}{M(M-1)}\sum_{m \neq l} \mathbf z^T_m \widehat{\Sigma}_m \mathbf z_l.
\]
Analogously to the homoskedastic case --- see equation \eqref{eq:homo_diagoff} --- the terms corresponding to $m=l$ are not taken into account in $\widehat S_2$, avoiding the introduction of potential biases from quadratic forms. Remark also that
\[
\widehat S_2 =  \frac{1}{M-1}\left[M\widehat{\boldsymbol{\nu}}^T \widehat{\boldsymbol{\mu}}
-  \widehat{S}_1\right],
\quad \text{ where } \quad
\widehat{\boldsymbol{\nu}} = \frac{1}{M}\sum_{m=1}^M \widehat{\Sigma}_m \mathbf z_m,
\]
 which is algorithmically be more convenient to compute. Also, $\widehat S_{NHe} = \widehat{\boldsymbol{\nu}}^T \widehat{\boldsymbol{\mu}}$. The estimates $\widehat S_{NHe}$ and $\widehat{S}_2$ have some similarities with $\widehat{\boldsymbol{\mu}}^T \widehat{\boldsymbol{\mu}}$ and $\widehat{\boldsymbol{\mu}}^T \widehat{\boldsymbol{\mu}}-(1/M)\Tr{C}$ as estimated in the homoskedastic case, see section \ref{subsubsec: muTmu}. However, $\mathbf z_m$ still interacts with the covariance matrix estimate within $\widehat{\boldsymbol{\nu}}$. This motivates a third estimate,
\[
\widehat S_3 = \frac{(M-3)!}{M!}\sum_{(m,l,k)\in A_M^3} \mathbf z^T_m \widehat{\Sigma}_k \mathbf z_l,
\]
where $A_M^3$ is the set of $3$-permutations of $M$. 
Looking at equation \eqref{eq:invest_intuition}, $\widehat{S}_3$ can also be obtained by replacing $\Sigma$ by a single estimate consisting of the average of the estimate of each model, where terms corresponding to $m=l$, $m=k$, or $l=k$ are ignored to avoid additional biases. To compute efficiently $\widehat{S}_3$, it can be rewritten as
\[
\widehat S_3  = \frac{1}{(M-1)(M-2)}
\left[M^2 \widehat{\boldsymbol{\mu}}^T \widehat U \widehat{\boldsymbol{\mu}} - M \widehat{V} - 2(M-1)\widehat S_2\right],
\]
with
\[
\widehat U = \frac{1}{M}\sum_{m=1}^M  \widehat{\Sigma}_m
\quad \text{ and }\quad
\widehat  V = \frac{1}{M}\sum_{m=1}^M \mathbf z^T_m  \widehat U \mathbf z_m.
\]
Finally, the proposed heteroskedastic estimates of ELM ensemble variance, noted $\hat \sigma^2_{S1}$, $\hat \sigma^2_{S2}$, $\hat \sigma^2_{S3}$ and $\hat \sigma^2_{NHe}$, are given by respectively adding $(1/M) \EstimVarf$ to $\widehat S_1$, $\widehat S_2$, $\widehat S_3$ and $\widehat S_{NHe}$.

To increase computation speed, approximated versions of $\widehat{S}_1, \widehat{S}_2$, $\widehat{S}_3$, and $\widehat{S}_{NHe}$ can be  obtained by replacing $\widehat{\Sigma}_m$ by $\widehat{\Omega}_m$ in the above reasoning. As a matter of fact, these two matrices are very close for sufficiently large $n$, but $\widehat{\Omega}_m$ is a diagonal matrix, while $\widehat{\Sigma}_m$ is a full matrix. Remark that the approximated version of $\hat \sigma^2_{NHe}$ is exactly the heteroskedastic estimate proposed in \cite{guignard2020model}.

$\quad$ 

\section{Synthetic experiments}

This section discusses the results obtained over different experimental settings. 
First, a simple non-regularized homoskedastic one-dimensional experiment is conducted. The variance estimate is thoroughly examined and assessed with quantitative measures and visualizations.  
Subsequently, the results are generalized to multi-dimensional settings with homoskedastic or heteroskedastic noise, both for the regularized and non-regularized cases.
Finally, CI estimation is discussed.
All the experiments presented will adopt the sigmoid as activation function, while input weights and biases will always be drawn uniformly between $-1$ and $1$.
All computations are done with the provided Python library, see the software availability at the end of the paper for more details.

\subsection{One-dimensional case}
\label{subsec:1Dcase}
To assess operationally the estimates proposed in section \ref{sec:estimate}, a simple one-dimensional simulated case study of $n=60$ training points is firstly proposed.
A trapeze shape probability density function defined by 
\[
\rho(x) = -\frac{x}{4\pi^2} + \frac{3}{4\pi}, \quad \text{ if } x\in [0, 2\pi],
\]
$\rho(x) = 0$ otherwise, is used to draw the input, i.e. the number of data decreases as $x$ increases. 
Outputs are generated according to
\[
y = \sin(x) + \varepsilon, 
\]
where $\varepsilon$ is an independent uniform noise $]-\sqrt{0.3}, \sqrt{0.3}[$ of constant variance $\sigma_\varepsilon^2 = 0.1$.

\begin{figure}
\begin{center}
\includegraphics[width=.475\textwidth]{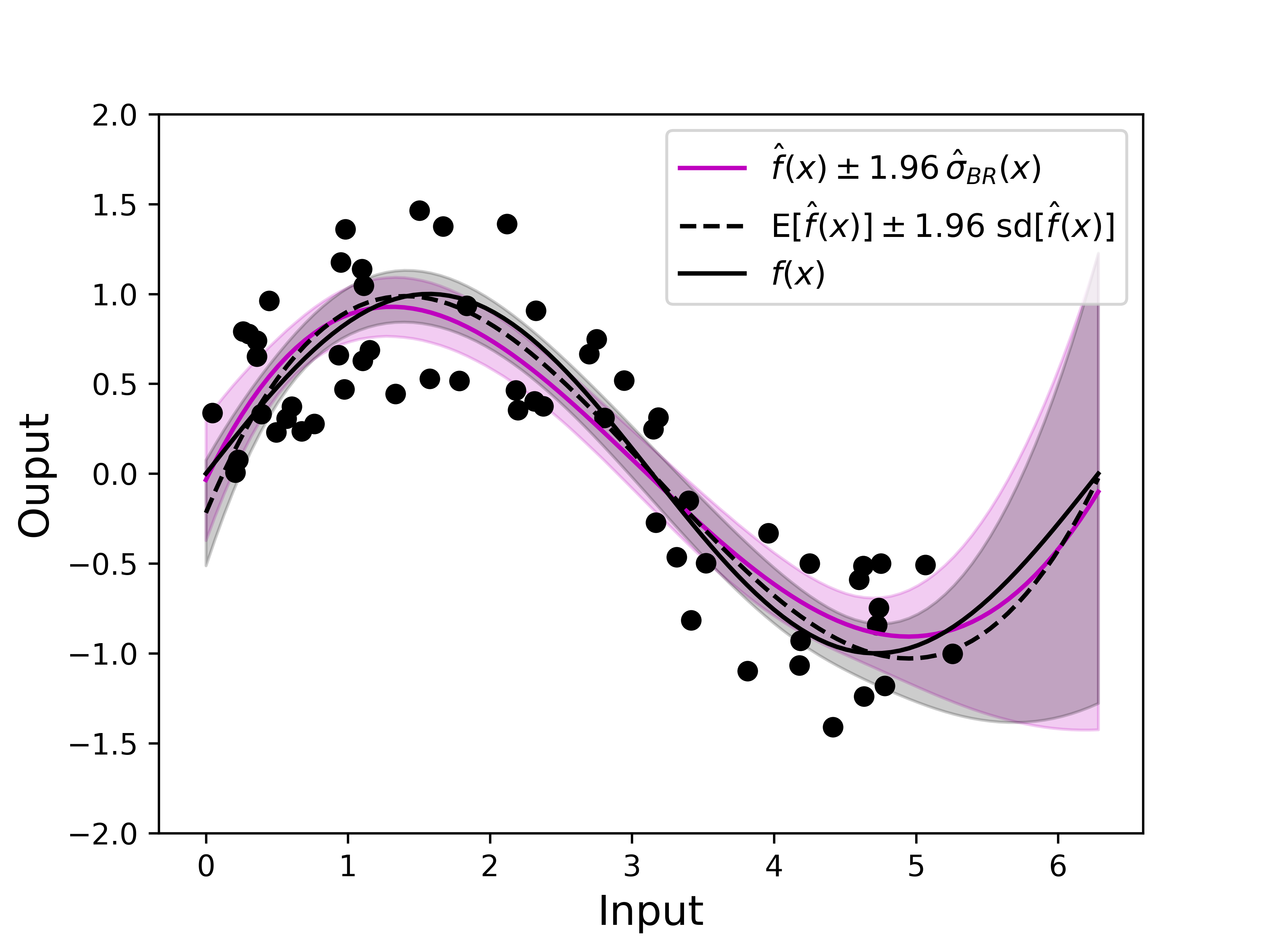}
\includegraphics[width=.475\textwidth]{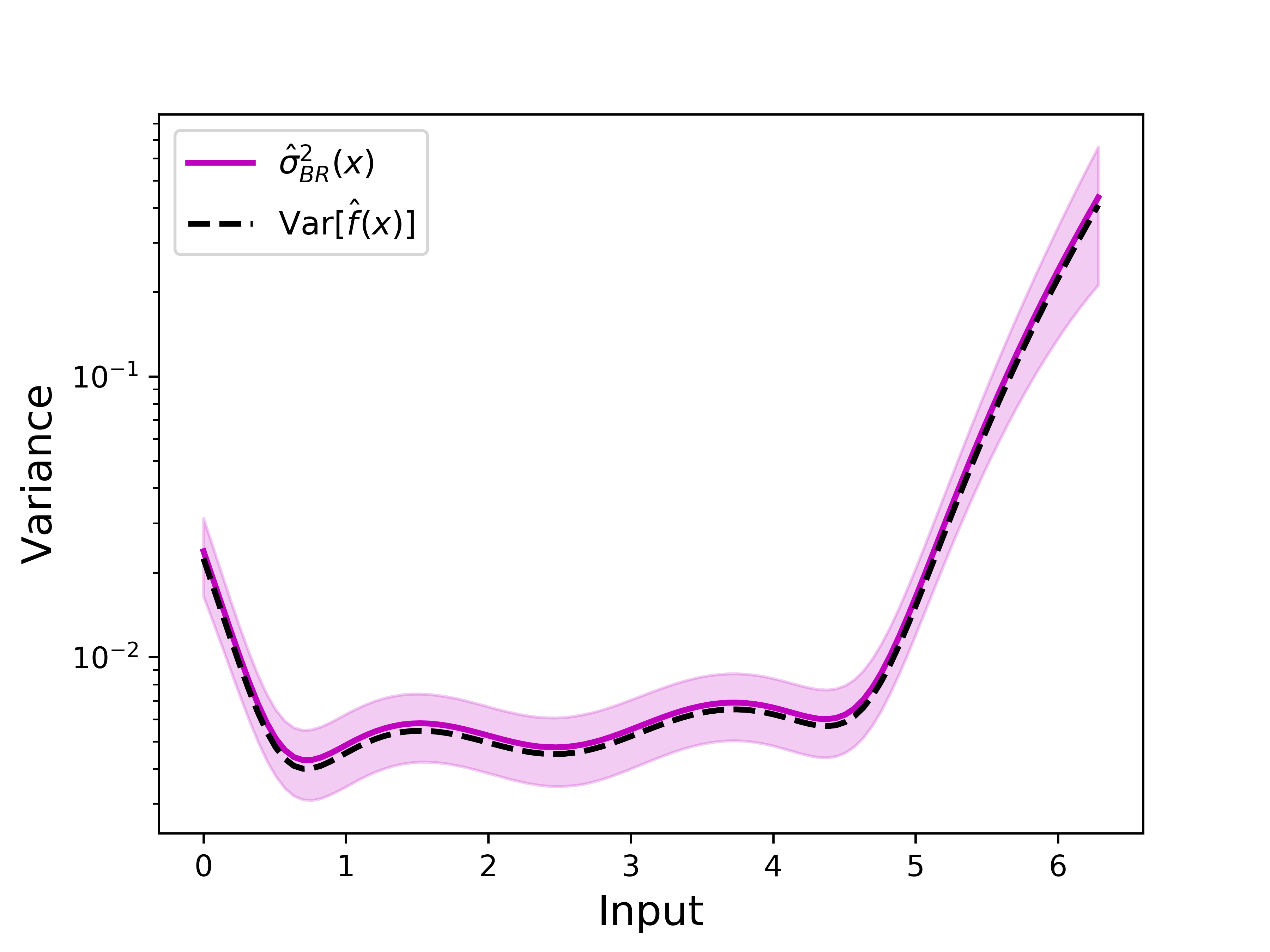}
\end{center}
\caption{One-dimensional synthetic experiment : (left) In magenta, a single estimation of $\hat f(x)$ with $M=10$ and its estimated $\pm1.96$ standard-error bands based on $\hat \sigma^2_{BR}$. In black dashed line, the mean of 10'000 ensembles for $M=10,$ with $\pm1.96$ standard-error bands. The true $f(x)$ is displayed in full black line; (right)  In magenta, the mean of $\hat \sigma^2_{BR}$ for $M=10$ with $\pm1.96$ standard-error bands, based on 1'000 replications of the experiment. In black dashed line,  the variance computed from the 10'000 ensembles, considered as ground truth. Note the logarithmic scale on the y-axis. 
}
 \label{fig1}
 \end{figure}

ELM ensembles are trained with $M = 5, 10, 20, 100$, allowing variance estimation. This experience is repeated 1'000 times. Each time, new outputs and new weights are drawn, but inputs are fixed. In order to avoid variability induced by hyper-parameter selection, a fixed number of neurons $N=4$ was chosen by a 5-fold cross validation process repeated 5 times on 1'000 dataset generations.
An example of one prediction is displayed in Figure \ref{fig1} (left) for $M = 10$. Estimation of pointwise standard-error bands based on $\pm 1.96 \, \hat \sigma_{BR}$ is also reported.

Although simulated datasets are produced by user controlled processes, the true value of the variance of $\hat f(x)$ remains unknown. In order to evaluate the estimate, $10'000$ ensembles with $M=5, 10, 20, 100$ and $N=4$ are trained with new outputs. The empirical mean and standard deviation of the $10'000$ ensembles is relatively close to respectively $\mathbb E[\hat f(x)]$ and  $\textnormal{sd}[\hat f(x)]=(\textnormal{Var}[\hat f(x)])^{1/2}$, and is reported as such for $M=10$ in Figure \ref{fig1}. In particular, the empirical variance of the $10'000$ ensembles will provide a reliable baseline for the variance estimation assessment and will be referred as the ground truth variance.

Figure \ref{fig1} (right) shows the mean  of $\hat \sigma^2_{BR}$ across the $1'000$ experiments with $\pm 1.96$ standard-error band. Compared with the ground truth variance,  the proposed estimate recovers effectively --- in average --- the variance from the $10'000$ simulations baseline. 
The  increasing variance in the borders due to the side effect of the modelling is fairly replicated. The uncertainty due to the trapezoidal shape of the input data distribution is also captured. Qualitatively, all aspects of the expected variance behaviour are globally reproduced. The naive estimate $\hat \sigma^2_{NHo}$ gives very similar results in one dimension, and is not shown. The improvement due to  considering the bias of  $\widehat{\boldsymbol{\mu}}^T  \widehat{\boldsymbol{\mu}}$ will be more relevant in the multi-dimensional case. 
However, one can still observe a residual bias for $\hat \sigma^2_{BR}$, partially due to the bias of $\hat \sigma^2_{\varepsilon}$ and to the  dependence between
$\hat \sigma^2_{\varepsilon}$ and the estimate of $\boldsymbol{\mu}^T \boldsymbol{\mu}$  --- see equations \eqref{eq:HomoNoiseBias} and \eqref{eq:final bias}.

\begin{table}
  \centering
  \resizebox{0.9\textwidth}{!}{%
  \begin{tabular}{cc|ccc|ccc}
      &  &  &   $n=60$ &   & &  $n=100$ &    \\
    
    \midrule
    \midrule
    
     \underline{$M=5$} &   
     & \underline{BR} & \underline{S3} & \underline{Grnd tr.} 
     & \underline{BR} & \underline{S3} & \underline{Grnd tr.} \\
     & $se_k$ 
     & 0.077 (0.006) & 0.080 (0.007) &  0.075 (---) 
     & 0.058 (0.003) & 0.060 (0.004) &  0.058 (---) \\
     Training set & $e_k$     
     & 0.005 (0.004) & 0.009 (0.005) &  --- 
     & 0.003 (0.002) & 0.005 (0.002) &  ---    \\
     & $re_k$   
     & 0.066 (0.052) & 0.113 (0.065) &   ---  
     & 0.046 (0.034) & 0.077 (0.039) &   --- \\
  
    \midrule
    
     & $se_k$ 
     & 0.077 (0.006) & 0.081 (0.007) &  0.075 (---) 
     & 0.060 (0.003) & 0.062 (0.004) &  0.059 (---)\\
     Testing set & $e_k$     
     & 0.005 (0.004) & 0.009 (0.005) &   ---   
     & 0.003 (0.002) & 0.005 (0.002) &   ---   \\
     & $re_k$     
     & 0.068 (0.053) & 0.115 (0.066) &  ---   
     & 0.044 (0.033) & 0.080 (0.039) &  ---\\
     
     \midrule
     \midrule

     \underline{$M=10$} &   
     & \underline{BR} & \underline{S3} & \underline{Grnd tr.} 
     & \underline{BR} & \underline{S3} & \underline{Grnd tr.} \\
     & $se_k$ 
     & 0.076 (0.005) & 0.079 (0.006) &  0.074 (---) 
     & 0.058 (0.003) & 0.060 (0.004) &  0.057 (---) \\
     Training set & $e_k$     
     & 0.005 (0.003) & 0.008 (0.004) &  --- 
     & 0.003 (0.002) & 0.004 (0.002) &  ---    \\
     & $re_k$   
     & 0.062 (0.046) & 0.109 (0.057) &   ---  
     & 0.044 (0.032) & 0.075 (0.037) &   --- \\
  
    \midrule
    
     & $se_k$ 
     & 0.076 (0.005) & 0.080 (0.006) &  0.074 (---) 
     & 0.060 (0.003) & 0.062 (0.004) &  0.059 (---)\\
     Testing set & $e_k$     
     & 0.005 (0.003) & 0.009 (0.004) &   ---   
     & 0.003 (0.002) & 0.005 (0.002) &   ---   \\
     & $re_k$     
     & 0.062 (0.046) & 0.111 (0.056) &  ---   
     & 0.043 (0.032) & 0.078 (0.037) &  ---\\
     
     \midrule
     \midrule

     \underline{$M=20$} &   
     & \underline{BR} & \underline{S3} & \underline{Grnd tr.} 
     & \underline{BR} & \underline{S3} & \underline{Grnd tr.} \\
     & $se_k$ 
     & 0.076 (0.005) & 0.079 (0.006) &  0.074 (---) 
     & 0.058 (0.003) & 0.060 (0.003) &  0.057 (---) \\
     Training set & $e_k$     
     & 0.005 (0.003) & 0.008 (0.004) &  --- 
     & 0.003 (0.002) & 0.004 (0.002) &  ---    \\
     & $re_k$   
     & 0.061 (0.045) & 0.109 (0.056) &   ---  
     & 0.042 (0.031) & 0.074 (0.036) &   --- \\

    \midrule
    
     & $se_k$ 
     & 0.076 (0.005) & 0.080 (0.006) &  0.074 (---) 
     & 0.059 (0.003) & 0.061 (0.004) &  0.059 (---)\\
     Testing set & $e_k$     
     & 0.005 (0.003) & 0.008 (0.004) &   ---   
     & 0.003 (0.002) & 0.005 (0.002) &   ---   \\
     & $re_k$     
     & 0.061 (0.044) & 0.110 (0.056) &  ---   
     & 0.042 (0.031) & 0.077 (0.037) &  ---\\
     
     \midrule
     \midrule

     \underline{$M=100$} &   
     & \underline{BR} & \underline{S3} & \underline{Grnd tr.} 
     & \underline{BR} & \underline{S3} & \underline{Grnd tr.} \\
     & $se_k$ 
     & 0.076 (0.005) & 0.079 (0.006) &  0.074 (---) 
     & 0.058 (0.003) & 0.060 (0.003) &  0.057 (---) \\
     Training set & $e_k$     
     & 0.004 (0.003) & 0.008 (0.004) &  --- 
     & 0.002 (0.002) & 0.004 (0.002) &  ---    \\
     & $re_k$   
     & 0.060 (0.045) & 0.109 (0.055) &   ---  
     & 0.041 (0.031) & 0.073 (0.036) &   --- \\
  
    \midrule
    
     & $se_k$ 
     & 0.076 (0.005) & 0.080 (0.006) &  0.074 (---) 
     & 0.059 (0.003) & 0.061 (0.004) &  0.059 (---) \\
     Testing set & $e_k$     
     & 0.004 (0.003) & 0.008 (0.004) &  --- 
     & 0.002 (0.002) & 0.005 (0.002) &  ---    \\
     & $re_k$   
     & 0.060 (0.044) & 0.110 (0.055) &   ---  
     & 0.041 (0.031) & 0.077 (0.036) &   --- \\
    \bottomrule
        \bottomrule
  \end{tabular}}
\caption{Results of the one-dimensional synthetic experiment.  Mean (standard deviation) of $se_k$, $e_k$ and $re_k$.}
\label{Results:A}
\end{table}

To assess quantitatively each estimation, a measure is needed between the true standard error of $\hat f (\mathbf x)$ and its estimations provided by each repetition of the experiment. Following \cite{Tibshirani}, let us look at the median of the $k^{th}$ standard error estimate over the training set,  
\[
se_k = \underset{1 \leq i \leq n}{\text{median}} \, \left(\hat \sigma_k(\mathbf x_i) \right),
\]
and the absolute error of the $k^{th}$ standard error estimate over the training set defined by
\[
e_k = \underset{1 \leq i \leq n}{\text{median}} \, |\hat \sigma_k(\mathbf x_i) - \sigma(\mathbf x_i)|,
\]
where $\hat \sigma^2_k(\mathbf{x}_i)$ is an estimate of $\sigma^2(\mathbf{x}_i)=\VarSmall{\hat f (\mathbf x_i) }$,  for $k = 1, \dots, 1'000$.
Also, the relative error $re_k$ of the $k^{th}$ standard error estimate over the training set is defined by
\[
re_k = \underset{1 \leq i \leq n}{\text{median}} \, \frac{|\hat \sigma_k(\mathbf x_i) - \sigma(\mathbf x_i)|}{\sigma(\mathbf x_i)},
\]
for $k = 1, \dots, 1'000$.
Similar measures are defined on a  random testing set of $1'000$ points.
In order to compute these quantities, $ \sigma(\mathbf x_i)$ is replaced by the ground truth standard deviation.

The means and standard deviations of $se_k$, $e_k$ and $re_k$ over the 1'000 experiment repetitions are presented in Table \ref{Results:A}. 
For the training set, the median of the ground truth standard error is recovered by the median of $\hat\sigma_{BR}$, judging through the  $se_k$ measure. Moreover, the mean and standard deviation of $e_k$ appear quite small.
The relative errors allow a better interpretation by comparing point-wise the absolute error with the true standard error. For instance for $M=10$, the mean of $re_k$ shows that --- on average --- the median error at training points represents $6.2\%$ of the true standard error. The results on the 1'000 testing points are similar, which shows that the estimation is good both at testing and  training points. 
Even for $M=5$, all error measures are quite satisfactory, as well as their standard deviations.
These results are also compared with $\hat \sigma^2_{S3}$. As expected for an homoskedastic dataset, estimate $\hat \sigma^2_{BR}$ based on homoskedastic assumption always results into better performance than $\hat \sigma^2_{S3}$ which is based on heteroskedastic assumption.

The same experiment is done with $n = 100$ and $N=5$. The results, reported in Table
\ref{Results:A}, show that all results are improved when increasing the number of data points, as expected.

\subsection{Multi-dimensional case}
\label{subsec:multivariate}

An example on a multi-dimensional case study is now investigated. Specifically, the synthetic dataset described by Friedman in \cite{Friedman} is considered, with fixed inputs $\mathbf x = (x_1, x_2, x_3, x_4, x_5)$ drawn independently from uniform distribution on the interval $[0,1]$ and outputs generated with an independent homoskedastic Gaussian noise according to
\begin{equation}\label{Friedman}
y(\mathbf x) = 10\sin(\pi x_1x_2) + 20(x_3-0.5)^2 + 10 x_4 + 5x_5 + \varepsilon, 
\end{equation}
with noise variance $\sigma_\varepsilon^2 = 0.5$. A number of $n=500$ training points are drawn. The number of neurons is chosen  by a similar cross-validation process as described above for the one-dimensional case, and fixed to $N=91$.
Ensembles are fitted and homoskedastic estimates $\hat \sigma^2_{BR}$ and $\hat \sigma^2_{NHo}$ are computed with $M=5, 10, 20, 100$. This is repeated $1'000$ times, while ground truth mean and variance are computed based on 10'000 ensembles, as in the previous experiment. The same experiment is conducted with  ensembles of regularized ELM. Tikhonov factor is selected with the help of generalized cross-validation \citep{golub1979generalized, piegorsch2015statistical} repeated on 1’000 dataset generations and set to $\alpha = 6 \cdot 10^{-6}$. Regularized version of $\hat \sigma^2_{BR}$ and $\hat \sigma^2_{NHo}$ are computed.

\begin{table}
  \centering
  \resizebox{0.9\textwidth}{!}{%
  \begin{tabular}{cc|ccc|ccc}
      &  &  &   Non-regularized &   & &  Regularized &    \\
    
    \midrule
    \midrule
    
     \underline{$M=5$} &   
     & \underline{BR} & \underline{NHo} & \underline{Grnd tr.} 
     & \underline{BR} & \underline{NHo} & \underline{Grnd tr.} \\
     & $se_k$ 
     & 0.273 (0.009) & 0.279 (0.009) &  0.255 (---) 
     & 0.250 (0.009) & 0.252 (0.009) &  0.222 (---) \\
     Training set & $e_k$     
     & 0.018 (0.008) & 0.023 (0.009) &  --- 
     & 0.025 (0.009) & 0.027 (0.009) &  ---    \\
     & $re_k$   
     & 0.070 (0.032) & 0.089 (0.037) &   ---  
     & 0.112 (0.039) & 0.121 (0.039) &   --- \\
  
    \midrule
    
     & $se_k$ 
     & 0.282 (0.010) & 0.289 (0.010) &  0.264 (---) 
     & 0.253 (0.009) & 0.255 (0.009) &  0.228 (---)\\
     Testing set & $e_k$     
     & 0.018 (0.008) & 0.024 (0.009) &   ---   
     & 0.025 (0.009) & 0.027 (0.009) &   ---   \\
     & $re_k$     
     & 0.069 (0.030) & 0.090 (0.034) &  ---   
     & 0.110 (0.039) & 0.119 (0.039) &  ---\\
     
     \midrule
     \midrule

     \underline{$M=10$} &   
     & \underline{BR} & \underline{NHo} & \underline{Grnd tr.} 
     & \underline{BR} & \underline{NHo} & \underline{Grnd tr.} \\
     & $se_k$ 
     & 0.266 (0.009) & 0.269 (0.009) &  0.247 (---) 
     & 0.243 (0.008) & 0.244 (0.008) &  0.214 (---) \\
     Training set & $e_k$     
     & 0.019 (0.008) & 0.022 (0.008) &  --- 
     & 0.028 (0.008) & 0.029 (0.008) &  ---    \\
     & $re_k$   
     & 0.078 (0.033) & 0.089 (0.034) &   ---  
     & 0.128 (0.035) & 0.132 (0.035) &   --- \\
  
    \midrule
    
     & $se_k$ 
     & 0.274 (0.009) & 0.277 (0.009) &  0.255 (---) 
     & 0.246 (0.008) & 0.247 (0.008) &  0.219 (---)\\
     Testing set & $e_k$     
     & 0.019 (0.008) & 0.023 (0.009) &   ---   
     & 0.028 (0.008) & 0.029 (0.008) &   ---   \\
     & $re_k$     
     & 0.078 (0.033) & 0.090 (0.034) &  ---   
     & 0.127 (0.035) & 0.132 (0.035) &  ---\\
     
     \midrule
     \midrule

     \underline{$M=20$} &   
     & \underline{BR} & \underline{NHo} & \underline{Grnd tr.} 
     & \underline{BR} & \underline{NHo} & \underline{Grnd tr.} \\
     & $se_k$ 
     & 0.263 (0.008) & 0.264 (0.008) &  0.243 (---) 
     & 0.239 (0.007) & 0.240 (0.007) &  0.210 (---) \\
     Training set & $e_k$     
     & 0.020 (0.008) & 0.021 (0.008) &  --- 
     & 0.028 (0.007) & 0.029 (0.007) &  ---    \\
     & $re_k$   
     & 0.082 (0.033) & 0.087 (0.033) &   ---  
     & 0.134 (0.034) & 0.137 (0.034) &   --- \\
  
    \midrule
    
     & $se_k$ 
     & 0.270 (0.008) & 0.272 (0.009) &  0.250 (---) 
     & 0.242 (0.007) & 0.243 (0.007) &  0.214 (---)\\
     Testing set & $e_k$     
     & 0.020 (0.008) & 0.022 (0.008) &   ---   
     & 0.028 (0.007) & 0.029 (0.007) &   ---   \\
     & $re_k$     
     & 0.081 (0.033) & 0.088 (0.033) &  ---   
     & 0.134 (0.034) & 0.137 (0.034) &  ---\\
     
     \midrule
     \midrule

     \underline{$M=100$} &   
     & \underline{BR} & \underline{NHo} & \underline{Grnd tr.} 
     & \underline{BR} & \underline{NHo} & \underline{Grnd tr.} \\
     & $se_k$ 
     & 0.259 (0.008) & 0.260 (0.008) &  0.239 (---) 
     & 0.236 (0.007) & 0.236 (0.007) &  0.207 (---) \\
     Training set & $e_k$     
     & 0.020 (0.008) & 0.020 (0.008) &  --- 
     & 0.029 (0.007) & 0.029 (0.007) &  ---    \\
     & $re_k$   
     & 0.084 (0.033) & 0.085 (0.033) &   ---  
     & 0.139 (0.033) & 0.139 (0.033) &   --- \\
  
    \midrule
    
     & $se_k$ 
     & 0.267 (0.008) & 0.267 (0.008) &  0.239 (---) 
     & 0.239 (0.007) & 0.239 (0.007) &  0.210 (---)\\
     Testing set & $e_k$     
     & 0.021 (0.008) & 0.021 (0.008) &   ---   
     & 0.029 (0.007) & 0.029 (0.007) &   ---   \\
     & $re_k$     
     & 0.084 (0.033) & 0.086 (0.033) &  ---   
     & 0.140 (0.033) & 0.140 (0.033) &  ---\\
    \bottomrule
        \bottomrule
  \end{tabular}}
\caption{Results of the multi-dimensional synthetic experiment with homoskedastic noise.  Mean (standard deviation) of $se_k$, $e_k$ and $re_k$.}
\label{Results:BHo500}
\end{table}

Results of the regularized and non-regularized versions of the experiment are reported in Table \ref{Results:BHo500}.
For both, the training $se_k$ tends to slightly overestimate the true standard deviation median over the training points. The testing $se_k$ has an analogous behaviour. Although the testing $se_k$ tends globally to be greater than the training $se_k$, the testing $e_k$ and $re_k$ are similar to the training $e_k$ and $re_k$. This suggests that regardless of the fact that the prediction is more uncertain at testing points, the variance estimation works at testing points as well as at the training points, as in the one-dimensional experiment. Note also that the true standard deviation median decreases as $M$ increases, as suggested by equation \eqref{equ:modvarensemblequad}. 
For the non-regularized case, 
the bias-reduced estimate $\hat \sigma^2_{BR}$ is systematically better than the $\hat \sigma^2_{NHo}$ estimate. 
Recalling that $\hat \sigma^2_{BR}$ reduce the bias by a quantity inversely proportional to $M$ --- see section \ref{subsec:homo} --- one observes that for $e_k$ and $re_k$ the improvement over $\hat \sigma^2_{NHo}$ is decreasing with $M$. 
The regularization mechanism increases the bias of the model while its variance decreases, which explains the decreasing of the true standard deviation median for a given $M$ from the non-regularized to the regularized case.
Moreover, for the regularized case, as the bias of the variance estimation depends directly from the conditional bias of the model, this could explain that the regularized experiment yields slightly weaker results in terms of $e_k$ and $re_k$. However, observe that $\hat \sigma^2_{BR}$ is still better than $\hat \sigma^2_{NHo}$.

To illustrate the heteroskedastic case, the same experiment is conducted with a non-constant noise variance. The Gaussian noise $\varepsilon(\mathbf{x})$ is now depending on the inputs variables through its variance by 
\[
\sigma_\varepsilon^2(\mathbf x) = 0.5  + 2\sin^2(\pi ||\mathbf x||_{\infty}),
\]
where $||\cdot||_\infty$ denotes the maximum norm.
The variance estimates $\hat \sigma^2_{S3}$, $\hat \sigma^2_{S2}$, $\hat \sigma^2_{NHe}$, $\hat \sigma^2_{S1}$ and $\hat \sigma^2_{BR}$ are computed in their approximated version $1'000$ times with $n=1000$, $M=5, 10, 20, 100$, and $N=109$.

\begin{figure}
\begin{center}
\includegraphics[width=.475\textwidth]{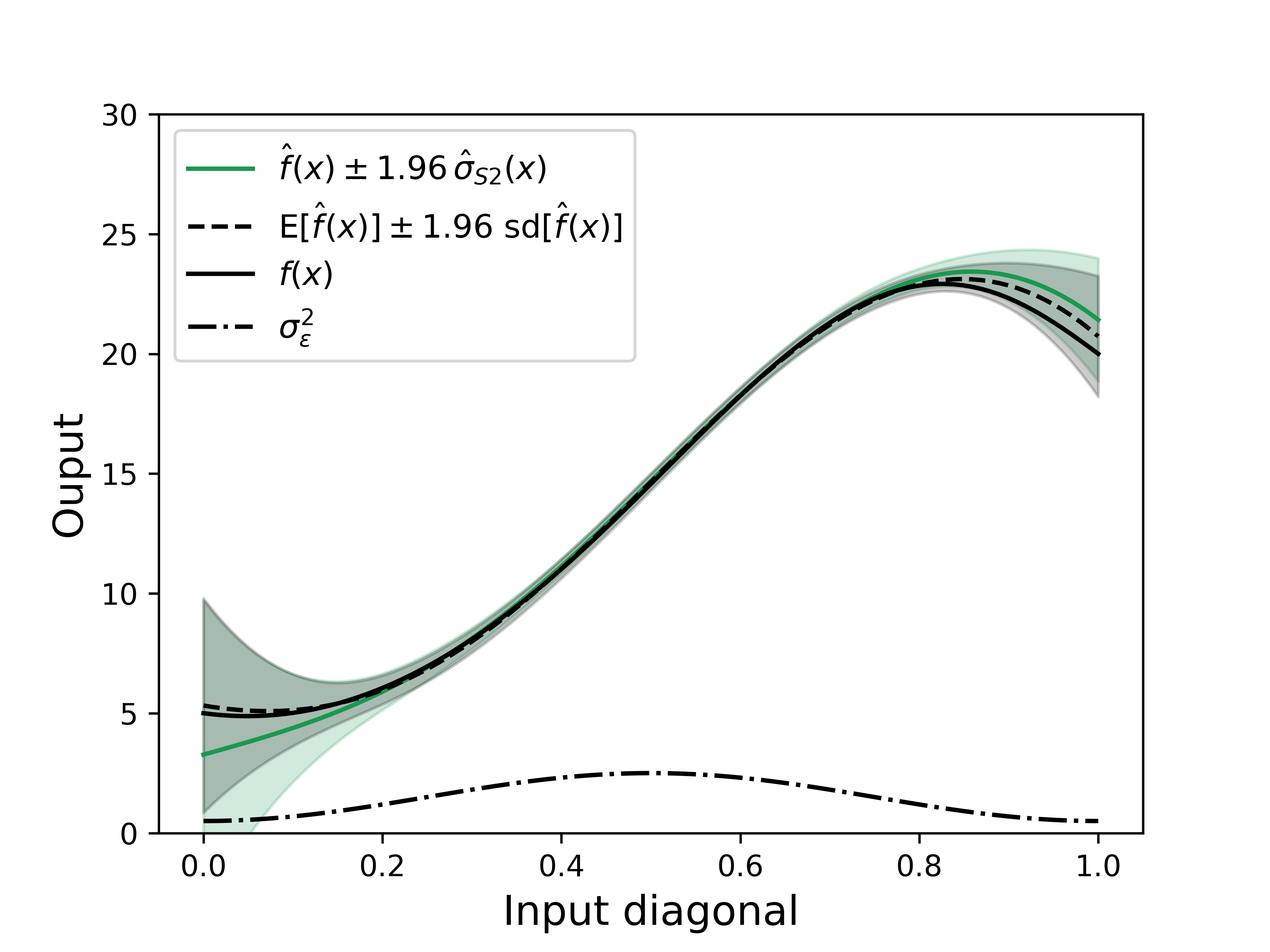}
\includegraphics[width=.475\textwidth]{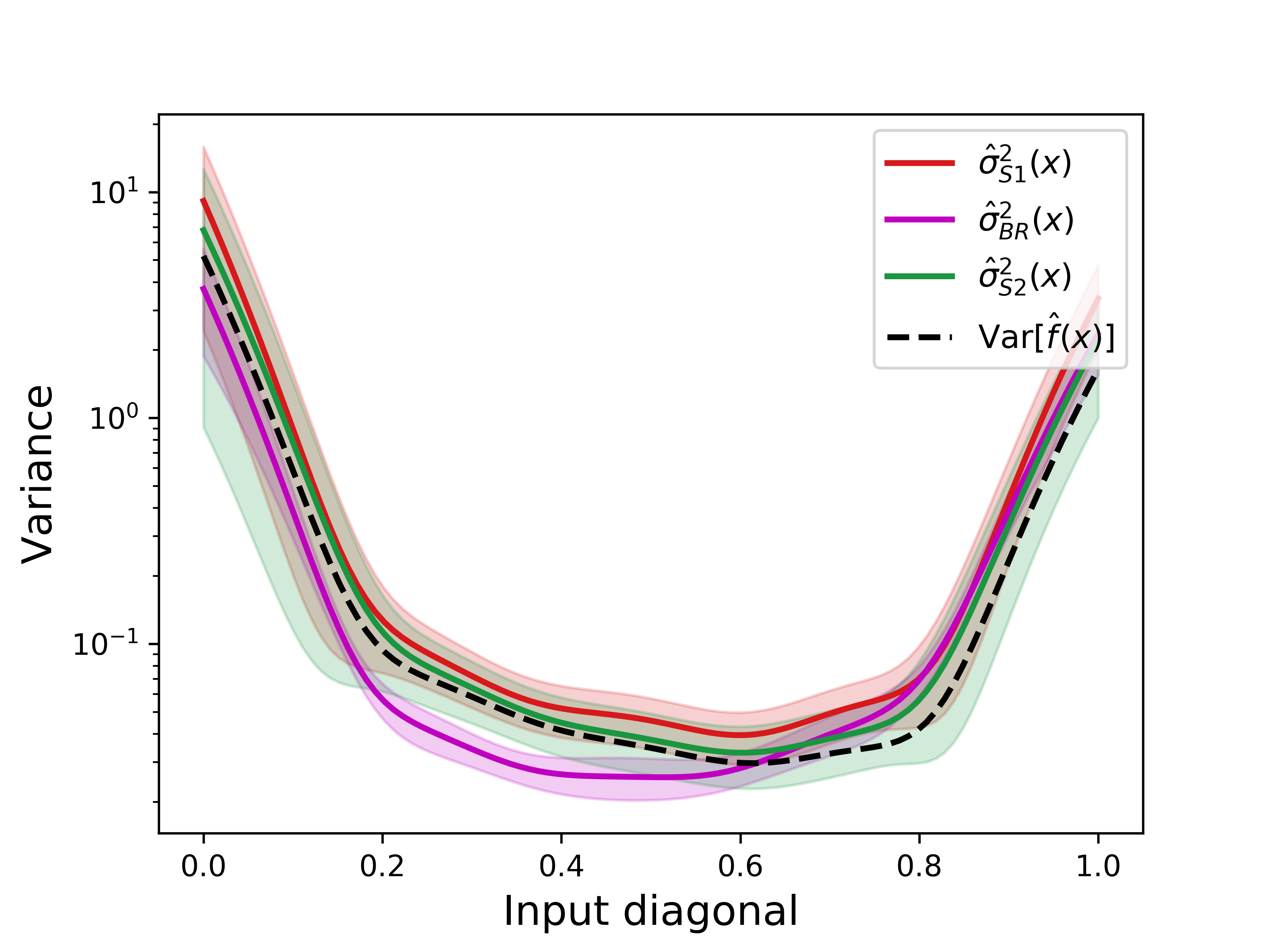}
\end{center}
\caption{Multi-dimensional heteroskedastic synthetic experiment : (left) In green, a single estimation of $\hat f(x)$ with $M=5$ and its estimated $\pm1.96$ standard-error bands based on $\hat \sigma^2_{S2}$. In black dashed line, the mean of 10'000 ensembles for $M=5,$ with $\pm1.96$ standard-error bands. The true $f(x)$ is displayed in full black line and the noise variance in dash-dotted black line; (right) The mean with $\pm1.96$ standard-error bands for $\hat \sigma^2_{S2}$ in green, $\hat \sigma^2_{S1}$ in red and $\hat \sigma^2_{BR}$ in purple, based on 1'000 replications of the experiment. In black dashed line, the variance computed from the 10'000 ensembles, considered as ground truth. The diagonal distance from the origin on the x-axis is measured in maximum norm.
}
 \label{fig2}
 \end{figure}

Although the results on the 5-dimensional hypercube input cannot be visualized, a small subset such as its diagonal can be plot, see Figure \ref{fig2}. On the left, prediction with $\pm1.96 \, \hat \sigma_{S2}$ is displayed for one experiment, for $M=5$. The true noise variance $\sigma_\varepsilon^2(\mathbf x)$ is also reported. On the right, results for $\hat \sigma^2_{S2}$, $\hat \sigma^2_{S1}$ and $\hat \sigma^2_{BR}$ for the 1'000 experiments are shown. Results for $\hat \sigma^2_{S3}$ and $\sigma^2_{Nhe}$ are visually close to $\hat \sigma^2_{S2}$ and are not reported. Heteroskedastic estimates reproduce fairly well the behaviour of the true variance, but $\hat \sigma^2_{S2}$ shows a smaller bias than $\hat \sigma^2_{S1}$ along the input diagonal. The homoskedastic estimates
$\hat \sigma^2_{BR}$ clearly fails to reproduce a coherent behaviour of the true variance, underestimating or overestimating it, depending on the location.

\begin{table}
  \centering
  \resizebox{0.9\textwidth}{!}{%
  \begin{tabular}{cc|cccccc}

    \midrule
    \midrule
    
     \underline{$M=5$} &   
     & \underline{S3} & \underline{S2} & \underline{NHe} 
     & \underline{S1} & \underline{BR} & \underline{Grnd tr.} \\
     & $se_k$ 
     & 0.279 (0.008) & 0.278 (0.007) &  0.285 (0.008) 
     & 0.308 (0.008) & 0.283 (0.007) &  0.253 (---) \\
     Training set & $e_k$     
     & 0.027 (0.005) & 0.027 (0.005) &  0.031 (0.006) 
     & 0.052 (0.008) & 0.046 (0.004) &   --- \\
     & $re_k$   
     & 0.109 (0.019) & 0.109 (0.019) &  0.125 (0.024) 
     & 0.211 (0.032) & 0.178 (0.015) &  --- \\
  
    \midrule
    
     & $se_k$ 
     & 0.302 (0.009) & 0.302 (0.009) &  0.309 (0.009) 
     & 0.335 (0.010) & 0.294 (0.008) &  0.275 (---) \\
     Testing set & $e_k$     
     & 0.029 (0.006) & 0.029 (0.006) &  0.035 (0.007) 
     & 0.060 (0.009) & 0.043 (0.004) &  --- \\
     & $re_k$     
     & 0.111 (0.022) & 0.111 (0.021) &  0.131 (0.027) 
     & 0.230 (0.034) & 0.157 (0.016) &  --- \\
     
     \midrule
     \midrule

     \underline{$M=10$} &   
     & \underline{S3} & \underline{S2} & \underline{NHe} 
     & \underline{S1} & \underline{BR} & \underline{Grnd tr.} \\
     & $se_k$ 
     & 0.273 (0.007) & 0.273 (0.007) &  0.276 (0.007) 
     & 0.303 (0.008) & 0.277 (0.007) &  0.247 (---) \\
     Training set & $e_k$     
     & 0.027 (0.005) & 0.027 (0.005) &  0.029 (0.005) 
     & 0.053 (0.008) & 0.047 (0.004) &  --- \\
     & $re_k$   
     & 0.111 (0.020) & 0.111 (0.020) &  0.120 (0.022) 
     & 0.224 (0.032) & 0.187 (0.015) &  --- \\
  
    \midrule
    
     & $se_k$ 
     & 0.296 (0.008) & 0.296 (0.008) &  0.299 (0.008) 
     & 0.330 (0.010) & 0.288 (0.008) &  0.267 (---) \\
     Testing set & $e_k$     
     & 0.030 (0.006) & 0.030 (0.006) &  0.032 (0.006) 
     & 0.062 (0.009) & 0.045 (0.004) &  --- \\
     & $re_k$     
     & 0.115 (0.022) & 0.114 (0.022) &  0.126 (0.025) 
     & 0.246 (0.034) & 0.168 (0.016) &  --- \\
     
     \midrule
     \midrule

     \underline{$M=20$} &   
     & \underline{S3} & \underline{S2} & \underline{NHe} 
     & \underline{S1} & \underline{BR} & \underline{Grnd tr.} \\
     & $se_k$ 
     & 0.270 (0.007) & 0.270 (0.007) &  0.272 (0.007) 
     & 0.301 (0.008) & 0.274 (0.007) &  0.244 (---) \\
     Training set & $e_k$     
     & 0.027 (0.005) & 0.027 (0.005) &  0.028 (0.005) 
     & 0.054 (0.007) & 0.048 (0.004) &  --- \\
     & $re_k$   
     & 0.113 (0.020) & 0.113 (0.020) &  0.118 (0.021) 
     & 0.230 (0.031) & 0.192 (0.015) &  --- \\
  
    \midrule
    
     & $se_k$ 
     & 0.292 (0.008) & 0.292 (0.008) &  0.294 (0.008) 
     & 0.327 (0.009) & 0.284 (0.007) &  0.263 (---) \\
     Testing set & $e_k$     
     & 0.030 (0.006) & 0.030 (0.006) &  0.031 (0.006) 
     & 0.063 (0.009) & 0.046 (0.004) &  --- \\
     & $re_k$     
     & 0.117 (0.023) & 0.117 (0.023) &  0.123 (0.024) 
     & 0.254 (0.033) & 0.173 (0.016) &  --- \\
     
     \midrule
     \midrule

     \underline{$M=100$} &   
     & \underline{S3} & \underline{S2} & \underline{NHe} 
     & \underline{S1} & \underline{BR} & \underline{Grnd tr.} \\
     & $se_k$ 
     & 0.268 (0.007) & 0.268 (0.007) &  0.268 (0.007) 
     & 0.299 (0.008) & 0.272 (0.007) &  0.242 (---) \\
     Training set & $e_k$     
     & 0.027 (0.005) & 0.027 (0.005) &  0.028 (0.005) 
     & 0.055 (0.007) & 0.049 (0.004) &  --- \\
     & $re_k$   
     & 0.115 (0.020) & 0.115 (0.020) &  0.116 (0.020) 
     & 0.236 (0.031) & 0.196 (0.015) &  --- \\
  
    \midrule
    
     & $se_k$ 
     & 0.290 (0.008) & 0.290 (0.008) &  0.290 (0.008) 
     & 0.325 (0.009) & 0.282 (0.007) &  0.261 (---) \\
     Testing set & $e_k$     
     & 0.030 (0.006) & 0.030 (0.006) &  0.030 (0.006) 
     & 0.064 (0.008) & 0.046 (0.004) &  --- \\
     & $re_k$     
     & 0.120 (0.023) & 0.120 (0.023) &  0.121 (0.023) 
     & 0.260 (0.033) & 0.178 (0.017) &  --- \\
    \bottomrule
        \bottomrule
  \end{tabular}}
\caption{Results of the multi-dimensional synthetic experiment with heteroskedastic noise.  Mean (standard deviation) of $se_k$, $e_k$ and $re_k$.}
\label{Results:BHe1000}
\end{table}

Quantitative results are shown in Table \ref{Results:BHe1000}. Again, the true standard deviation median decreases when $M$ increases, and the training (testing) $se_k$ tends to somewhat overestimate the true training (testing) standard deviation median. 
The homoskedastic estimate $\hat \sigma^2_{BR}$ no longer gives the best results because of the heteroskedastic nature of the data, which justify the use of heteroskedastic estimates. The estimate $\hat \sigma^2_{S1}$ --- which reuse HCCME in its original form --- gives the worst results as suspected in section \ref{subsec:hetero}. The naive heteroskedastic estimate $\hat \sigma^2_{NHe}$ gives in general reasonable results. However, the heteroskedastic estimates $\hat \sigma^2_{S2}$ --- which was developed based on the insight given by the homoskedastic case in section \ref{subsec:homo} --- allows to improve the results by a quantity decreasing with $M$, as expected. Finally, results from $\hat \sigma^2_{S3}$ are very close to $\hat \sigma^2_{S2}$.

Regularized versions of the heteroskedastic estimates were also investigated and  $\hat \sigma^2_{S2}$ and  $\hat \sigma^2_{S3}$ gave quite reasonable results too.

\subsection{Towards confidence intervals}

\begin{figure}
\begin{center}
\includegraphics[width=.320\textwidth]{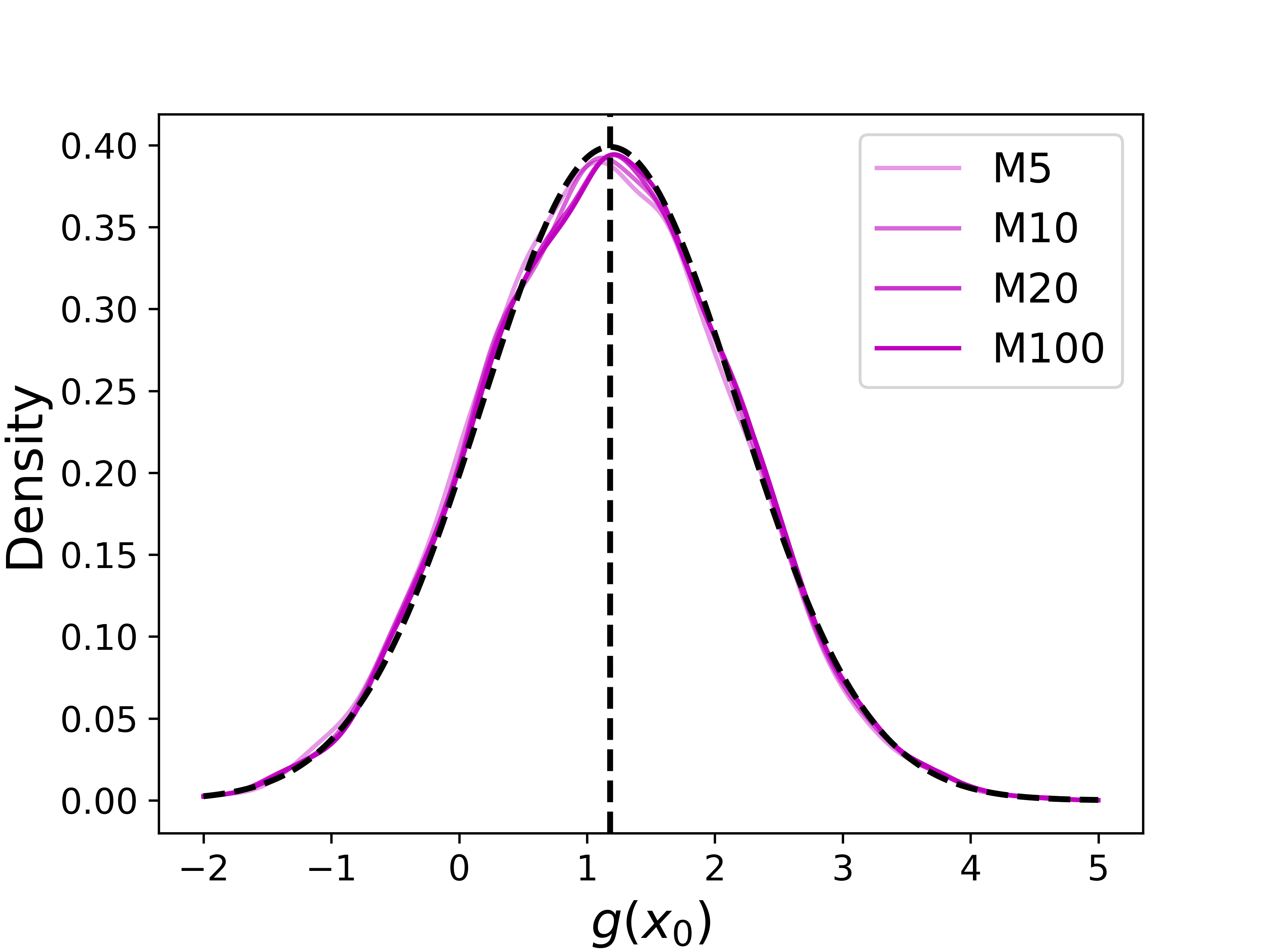}
\includegraphics[width=.320\textwidth]{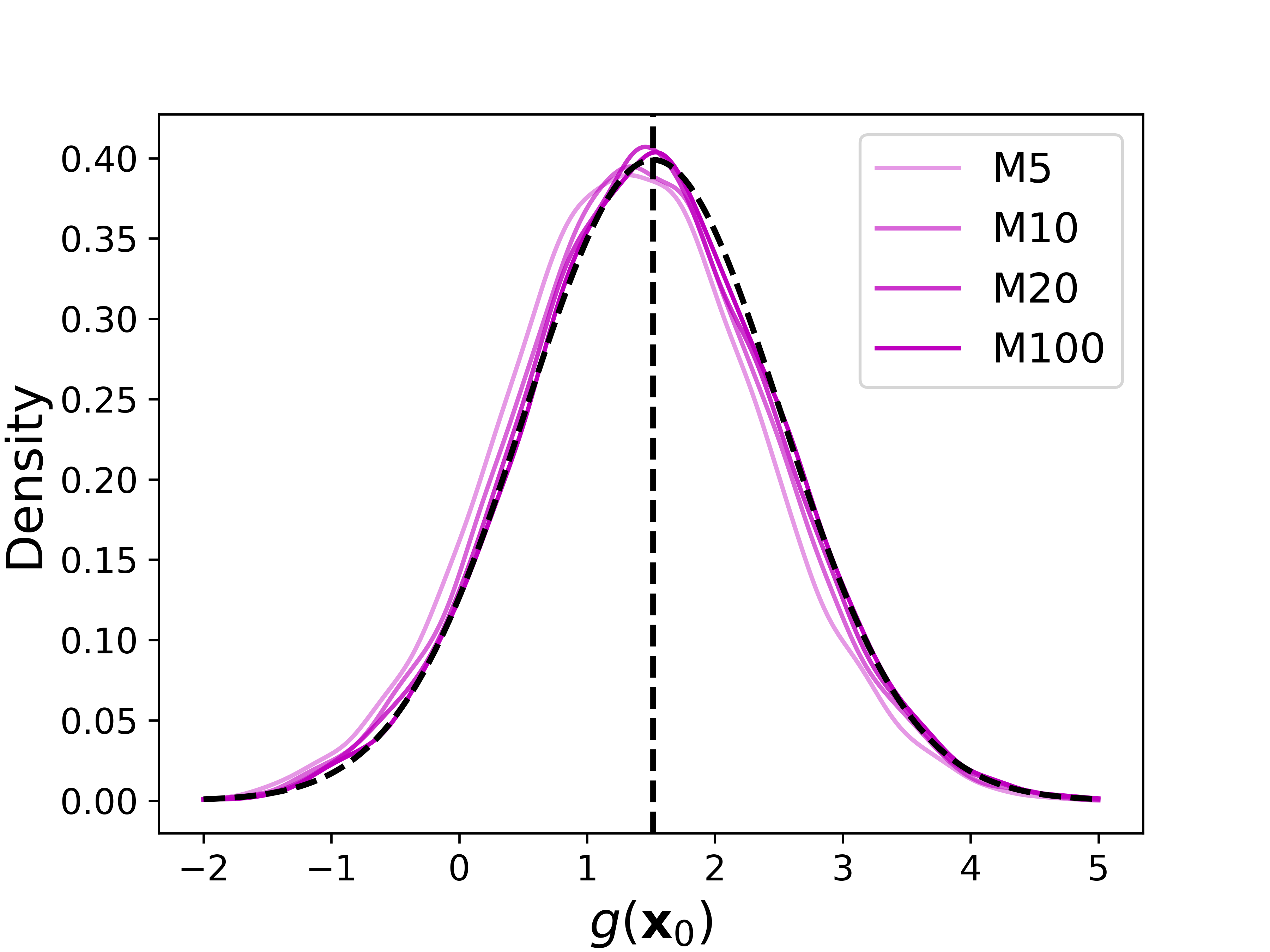}
\includegraphics[width=.320\textwidth]{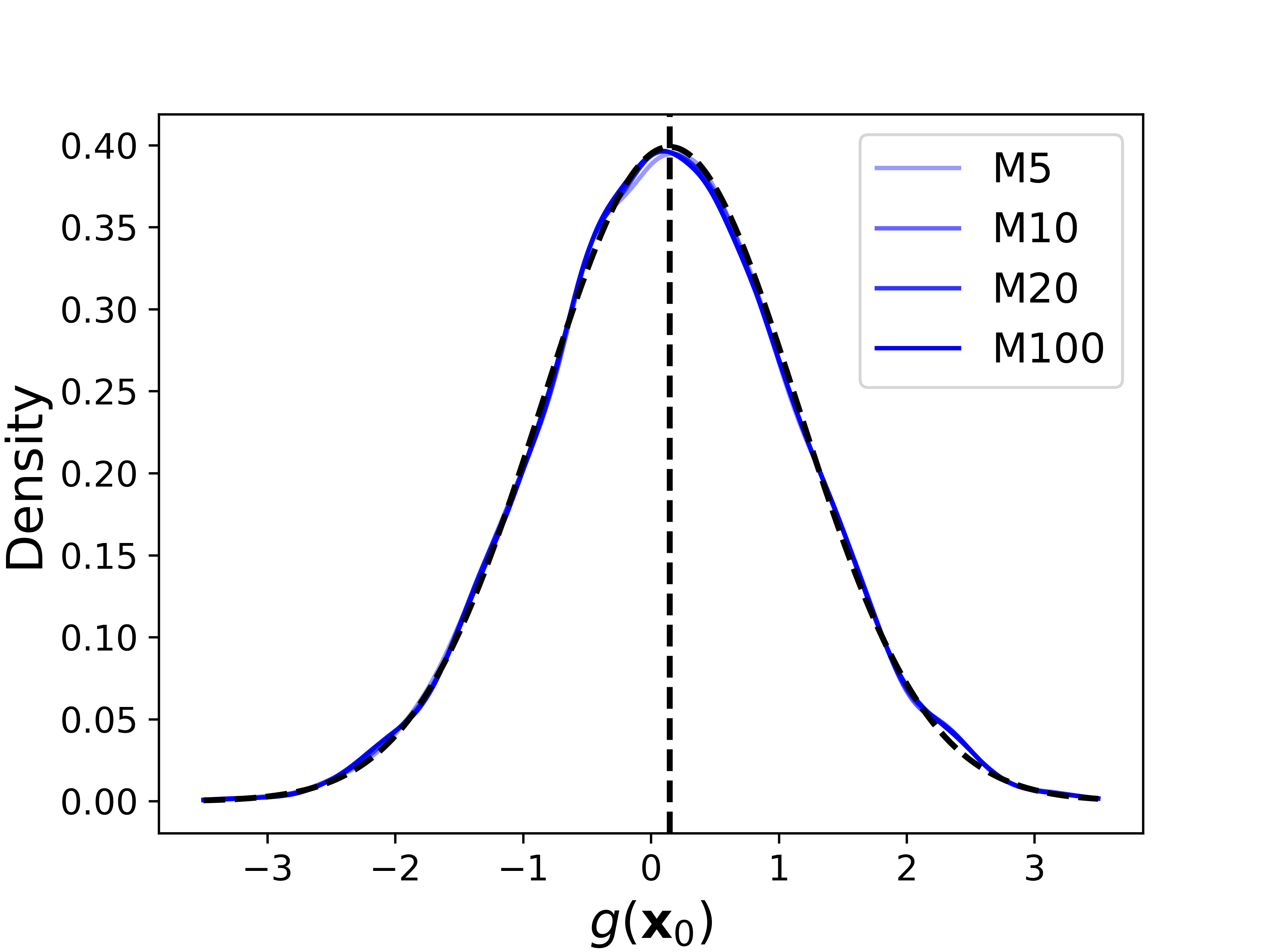}
\end{center}
\caption{Kernel density estimates of $g(\mathbf x_0)$; (left) for the one-dimensional case, at $x_0 = \pi/4$, (middle) for the multi-dimensional case with homoskedastic noise, at $\mathbf x_0 = 1/2 \cdot (1, 1, 1, 1, 1)$, (right) at the same point with regularized ELM ensemble of $N=300$ neurons.  In black dashed line, Gaussian distributions with unit variance and mean (vertical dashed line) determined by the ratio between the bias and the standard deviation of $\hat f(\mathbf x_0)$ 
}
 \label{fig3}
 \end{figure}

Although visually it is tempting to say so, there is so far no guarantee that $\hat f(x) \pm 1.96 \, \hat \sigma_{BR}(x)$ in Figure \ref{fig1} (left) or $\hat f(x) \pm 1.96 \cdot \hat \sigma_{S2}(x)$ in Figure \ref{fig2} (left) define a 95\% CI for $f(x)$.
Let us investigate the possibility to build (approximate) CI in particular cases.
The distribution of
\[
g(\mathbf x) = \frac{\hat f(\mathbf x) - f(\mathbf x)}{\text{sd}[\hat f(\mathbf x)]},
\]
is unknown.
However, for our simulated case studies, remark that it is very close to a Gaussian distribution. Kernel density estimates --- based on the 10'000 replications done in previous sections --- are shown in Figure \ref{fig3} at some example points at which $\hat f(\mathbf{x})$ exhibits significant bias. Figure \ref{fig3} (left) displays the distribution of $g(x_0)$ at $x_0 = \pi/4$ for the one-dmensional case. Figure \ref{fig3} (middle) visualizes the distribution of $g(\mathbf x_0)$ at $\mathbf x_0 = 1/2 \cdot (1, 1, 1, 1, 1)$ for the Friedman dataset with homoskedastic noise, which corresponds to the center of the 5-dimensional hypercube and the middle point of the input diagonal. Distribution of $g(\mathbf x_0)$ behaves similarly for the Friedman dataset with heteroskedastic noise (not shown) and other experiments were conducted with noise from Student laws showing the same behaviour. 
This suggests that for ELM ensembles, $g(\mathbf x)$ and $\hat f (\mathbf x)$ may asymptotically follow a Gaussian distribution. However, dependencies exist between the components of $\mathbf z_m$, and also between the members of the ELM ensemble. Therefore, the classical central limit theorem is not directly applicable, and it seems hard to straightforwardly conclude to Gaussianity in case of large sample size $n$ or large $M$. 
In spite of knowing if one can prove or disprove this conclusion, let us assume that the distribution of $g(\mathbf x)$ is (asymptotically) Gaussian in the remainder of this section. As a matter of fact, $g(\mathbf x)$ has a unit variance but it is not centred, due to the bias of $\hat f (\mathbf x)$. Its mean is given by 
\[
\E{g(\mathbf x)} = \frac{\text{Bias}[\hat f(\mathbf x)]}{\text{sd}[\hat f(\mathbf x)]},
\]
and is reported in Figure \ref{fig3} as a vertical dashed black line. Obviously, this quantity is unknown in practice
but necessary to build a reliable CI for $f(\mathbf x)$. However, if the bias of $\hat f(\mathbf x)$ is negligible relatively to its variance, then $g(\mathbf x)$ is close to centred, and approximate point-wise CI can be derived based solely on an estimation of the variance of $\hat f(\mathbf{x})$. That is, if $g(\mathbf x)$ is close to centred, then the estimated $\pm 1.96$ standard-error around $\hat f (\mathbf x)$ define an approximate point-wise 95\% CI for $f(\mathbf x)$.

 \begin{figure}
\begin{center}
\includegraphics[width=.320\textwidth]{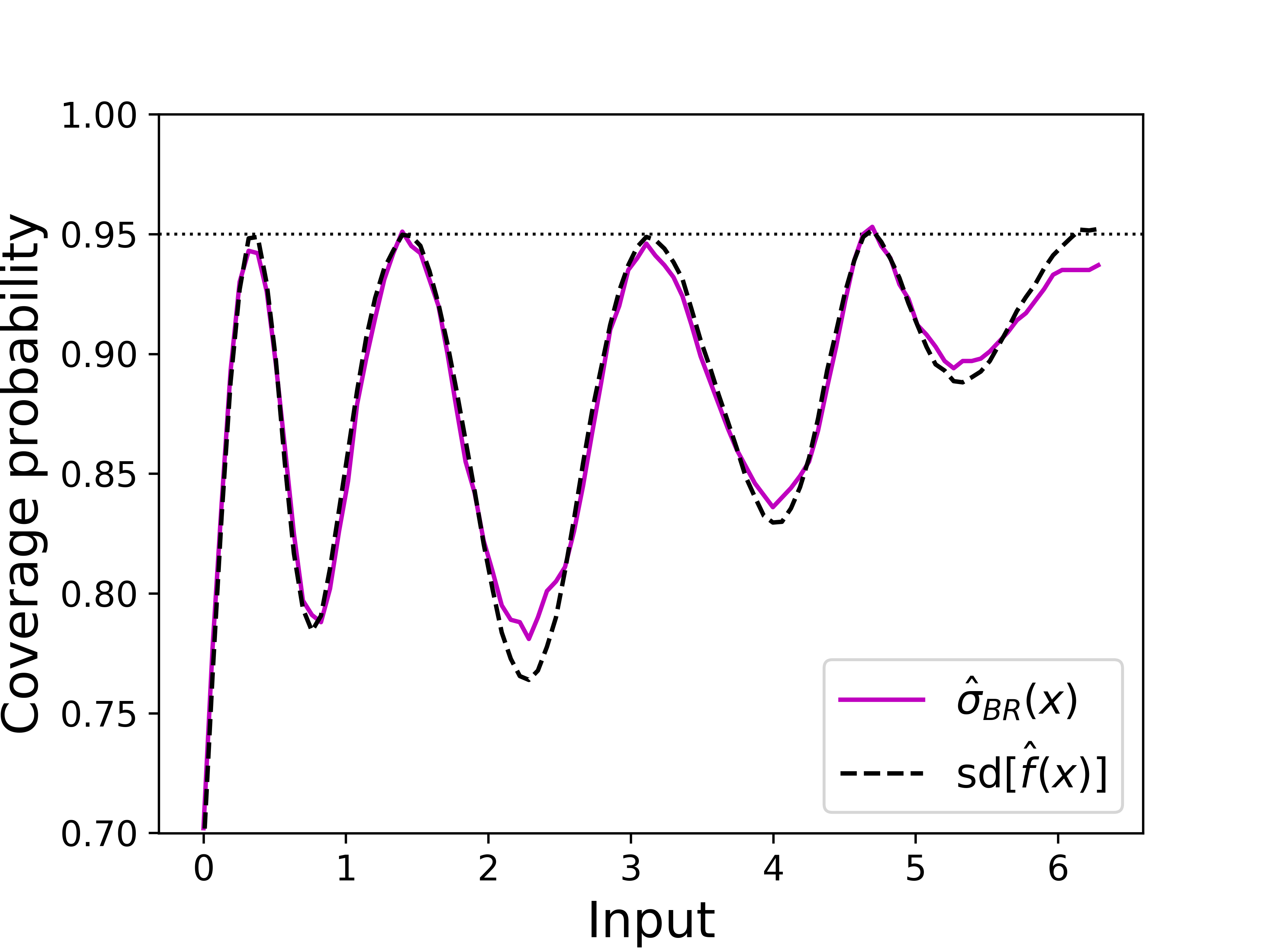}
\includegraphics[width=.320\textwidth]{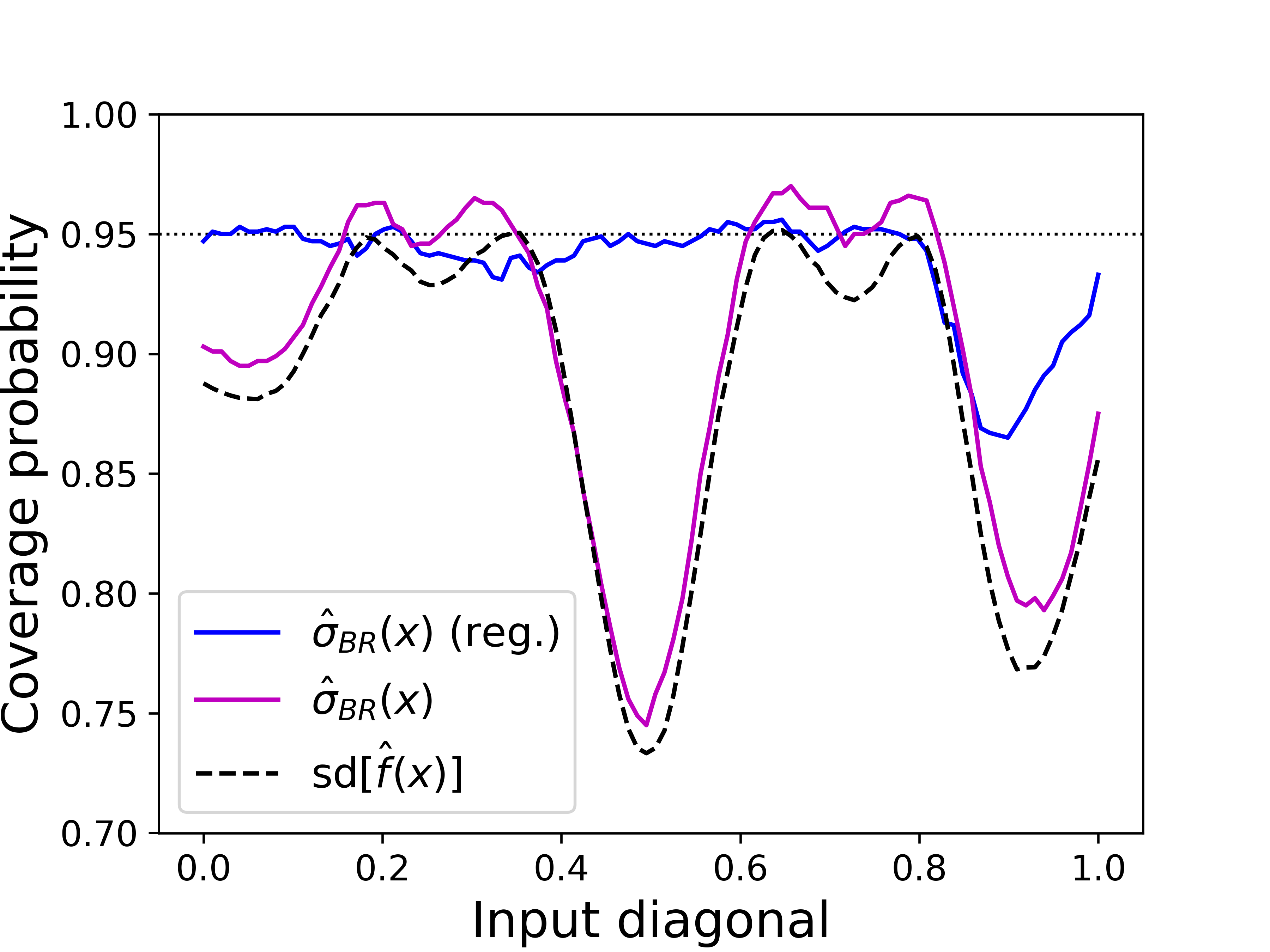}
\includegraphics[width=.320\textwidth]{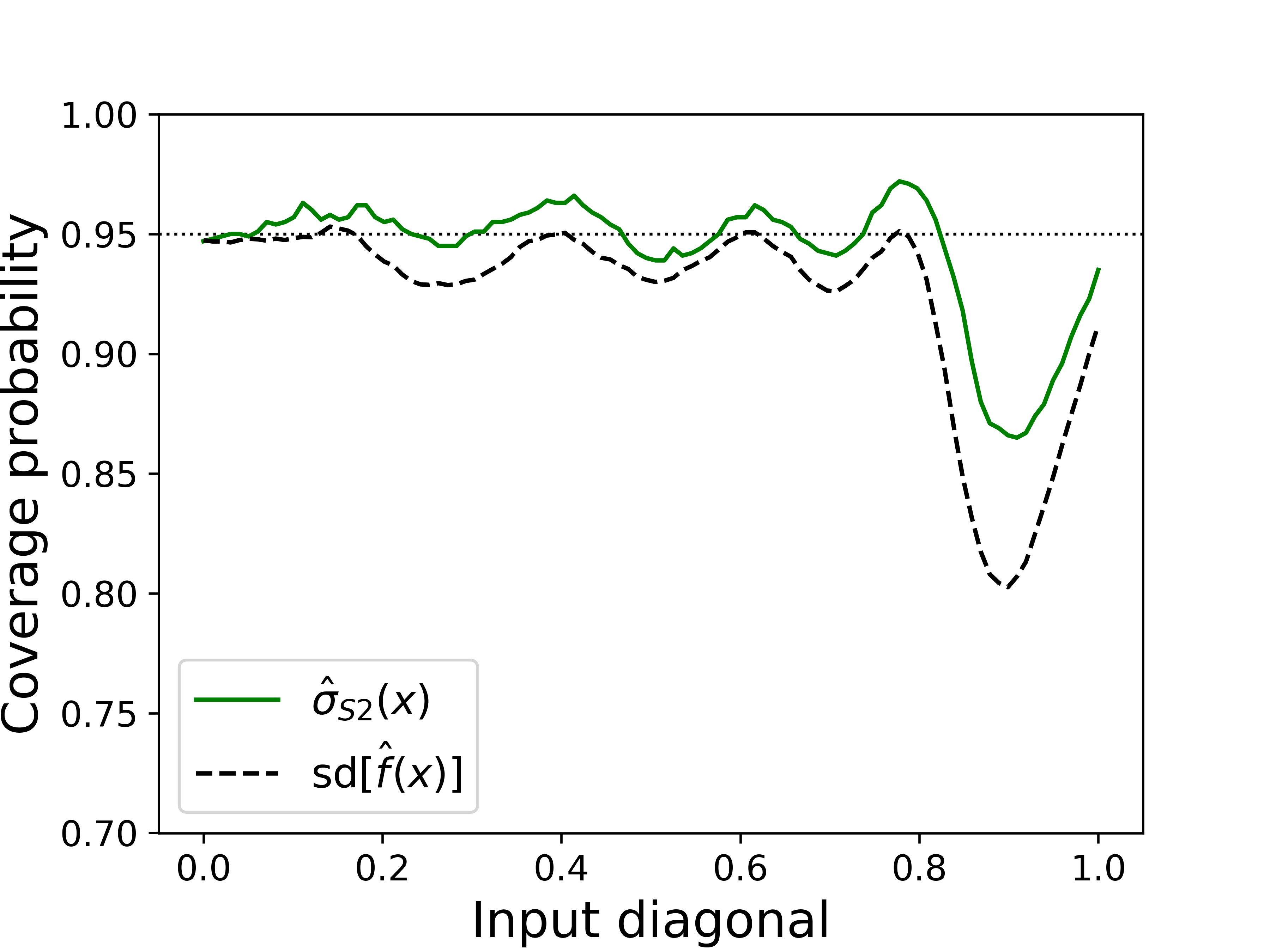}
\end{center}
\caption{Coverage probabilities for the one-dimensional case with $M = 10$ (left),  for the multi-dimensional case with $M=5$, with homoskedastic noise (middle), and with heteroskedastic noise (right).  The true coverage probability is fixed at 95 \% (dotted line). The actual coverage probability is reported for $\hat f(x)\pm 1.96 \, \hat \sigma_{BR}$ (in purple) and $\hat f(x)\pm 1.96 \, \hat \sigma_{S2}$ (in green). It is also shown for $\hat f(x)\pm 1.96 \, \text{sd}[\hat f(x)]$ in black dashed line, for comparison purpose. In the homoskedastic multi-dimensional case, the actual coverage probability is also reported for $\hat f(x)\pm 1.96 \, \hat \sigma_{BR}$ for ensemble of $M=5$ regularized ELM of $N=300$ neurons (in blue).
}
\label{fig4}
\end{figure}

Figure \ref{fig4} plots for some of the previous experiments the coverage probability of the approximate point-wise 95\% CI, i.e. the proportion of time that the estimated confidence interval actually contains the true $f(\mathbf x)$ among all the $1'000$ experiment repetitions.
Figure \ref{fig4} (left) shows the one-dimensional case for $M=10$.
The black dashed line indicates the proportion of time that $f(x)$ lies within $\hat f (x) \pm  1.96 \, \text{sd}[\hat f(x)]$, computed on the basis of the $10'000$ simulations baseline. 
Observe that around each point where the bias vanishes --- seen on Figure \ref{fig1} (left) --- this proportion comes closer to the true coverage probability of $0.95$ defined by the confidence level. 
Conversely, for instance at $x_0=\pi/4$ --- which is near the point with the smallest variance, see Figure \ref{fig1} (right) --- the bias is high relatively to the variance, then $g(\mathbf x_0)$ is far from centred, which implies a wrong construction of the CI, leading to a bad result. 
Obviously, $\text{sd}[\hat f(x)]$ is not available in practice, and looking at the actual coverage probability of the estimated CI $\hat f (x) \pm  1.96 \, \hat \sigma_{BR}(x)$ is more interesting. However, note that it reproduces quite fairly the same behaviour, as expected. 

The non-regularized multi-dimensional experiment for $M=5$ is also displayed in Figure \ref{fig4} for homoskedastic case, also estimated with $\hat f (x) \pm  1.96 \, \hat \sigma_{BR}(x)$ (middle), and for heteroskedsatic case, estimated with $\hat f (x) \pm  1.96 \hat \sigma_{S2}(x)$ (right). For both, the actual coverage probability is globaly greater than the proportion based on the theoretical CI build with the true $\text{sd}[\hat f(x)]$. This is partially explained by the overestimation of the $\hat f(\mathbf x)$ standard deviation --- see section \ref{subsec:multivariate}. In some low bias regions, this results in slightly conservative CI, i.e. the actual coverage probability is greater than the true coverage probability of 95\%. Observe that around $\mathbf x_0 = 1 / 2 \cdot (1, 1, 1, 1, 1)$ the actual coverage probability is especially bad for the homoskedastic case, while it is quite reasonable for the heteroskedastic case. This is explained by the fact that the heteroskedastic noise variance around the center of the hypercube is up to five times more than the homoskedastic variance, which implies an increase of the variance of $\hat f(\mathbf x)$ and a decrease of  $\mathbb E[g(\mathbf x)]$ around $\mathbf x_0$. Finally, the actual coverage probability is quite satisfying for the heteroskedastic case.

Clearly, the effectiveness of the CI estimation for $f(\mathbf x)$ is highly dependant on the dataset at hand and significant bias of $\hat f(\mathbf x)$ relatively to its variance can lead to highly permissive and bad CI for  $f(\mathbf x)$. However, it is possible to identify potential paths to overcome this problem. 
Firstly, note that even if the bias of $\hat f (\mathbf x)$ is too important to be ignored, the estimated $\pm 1.96$ standard-error bands around $\hat f(\mathbf x)$ still provides a reliable CI for $\mathbb E[\hat f(\mathbf x)]$. 
Secondly, the bias could be estimated.
Thirdly, a manner of reducing the bias is to smooth $\hat f(\mathbf{x})$ slightly less than what  would be appropriate \citep{hall1992effect}, for instance through regularization. 
For the latter, one provides here an example for the multi-dimensional case with homoskedastic noise. 

Ensemble of $M=5$ regularized ELM is trained with $N=300$. Selecting voluntarily a bigger number of neurons increases the model complexity, then reduces the model bias. But increasing complexity also implies increasing the model variability, which puts the model in an overfitting situation that the regularization mechanism controls at the expense of the introduction of an additional bias. The general cross-validation estimate results in a Tikhonov factor of $10^{-4}$ which introduces to much bias. Then, $\alpha = 10^{-6}$ is empirically set to decrease the amount of smoothing hence alleviating the bias, at the expense of an higher variance.
To measure predictive performance of the model, the mean squared error ($MSE$) and relative mean squared error ($RE$) are defined on the training set by
\[
MSE = \frac{1}{n}\sum_{i=1}^n \left(y_i - \hat f(\mathbf x_i) \right)^2
\quad \text{ and } \quad
RE = \frac{\frac{1}{n}\sum_{i=1}^n \left(y_i - \hat f(\mathbf x_i)\right)^2}{\frac{1}{n}\sum_{i=1}^n \left( y_i - \frac{1}{n}\sum_{i=j}^n y_j \right)^2},
\]
and similar measures are defined on the testing set.
Generally speaking, lower values of $MSE$ and $RE$ are better. A value higher than 1 for $RE$ indicates that the model performs worse than the mean \citep{golay2017feature}.
Note also that $RE$ can be interpreted as an estimation of the ratio between the residual variance and the data variance. 

Table \ref{Results:CI} shows the quantitative results
of the regularized model with $N=300$ compared to the non-regularized model done previously with $N=91$.
\begin{table}
  \centering
  \resizebox{0.65\textwidth}{!}{%
  \begin{tabular}{cc|cc|cc}
     &   & $N=91$ & Grnd tr. &  $N=300$ (reg.), & Grnd tr. \\
    \midrule
    \midrule
     & $MSE$   
     & 0.441 (0.030) & ---
     & 0.361 (0.025) & ---  \\
     & $RE$   
     & 0.017 (0.001) & ---
     & 0.014 (0.001) & ---  \\
     Training set & $se_k$ 
     & 0.273 (0.009) & 0.255 (---) 
     & 0.297 (0.010) & 0.294 (---)  \\
     & $e_k$     
     & 0.018 (0.008) & ---
     & 0.009 (0.006) & ---  \\
     & $re_k$   
     & 0.070 (0.032) & ---
     & 0.031 (0.022) & ---  \\
     
    \midrule
    
     & $MSE$   
     & 0.679 (0.037) & ---
     & 0.682 (0.041) & ---  \\    
     & $RE$   
     & 0.028 (0.002) & ---
     & 0.028 (0.002) & ---  \\
     Testing set & $se_k$ 
     & 0.282 (0.010) & 0.264 (---) 
     & 0.312 (0.011) & 0.307 (---)  \\
     & $e_k$     
     & 0.018 (0.008) & ---
     & 0.009 (0.007) & ---  \\
     & $re_k$   
     & 0.069 (0.030) & ---
     & 0.031 (0.022) & ---  \\

     \bottomrule
        \bottomrule
  \end{tabular}}
\caption{Comparison between non-regularized model with $N=91$ neurons and regularized model with $N=300$ and $\alpha = 10^{-6}$,  on the multi-dimensional synthetic experiment with homoskedastic noise. For both model, $M=5$ and the variance is estimated with $\hat \sigma^2_{BR}$.
Mean (standard deviation) of $MSE$, $se_k$, $e_k$ and $re_k$.}
\label{Results:CI}
\end{table}
In average among the $1'000$ experiments, the testing $MSE$ is slightly better for $N=91$.
However, the testing $RE$ shows that in both cases it represents $2.8\%$ of the data variance and no significant difference is identifiable.
As expected, the $se_k$ of the true variance --- and its estimation --- is greater for $N=300$. The variance is better estimated, as shown by $e_k$ and $re_k$. This is likely due to the model bias reduction, which probably implies a decrease of the bias of the noise estimation, and therefore of the bias of the variance estimates --- see section \ref{subsec:homo}.
Figure \ref{fig3} (right) visualizes the distribution of $g(\mathbf x_0)$ at the center of the 5-dimensional hypercube. Comparing with the first model $N=91$ --- Figure \ref{fig3} (middle) --- the distribution  of $g(\mathbf x_0)$ is far closer to a centered Gaussian. The coverage probability is also shown and compared in Figure \ref{fig4} (middle), where results are considerably improved, especially at $\mathbf x_0 = 1/2 \cdot (1, 1, 1, 1, 1)$. 
Summarizing, while the CI are then globally correctly estimated, the predictive performance are almost the same.


\section{Conclusion}
This paper discussed variance of (regularized) ELM under general hypothesis and its estimation through small ensembles of retrained ELMs under homoskedastic and heteroskedastic hypothesis. 
As ELM is nothing more than a linear regression in a random feature space, analytical results can be derived  by conditioning on the random input weights and biases. In particular, the variance of $\hat f(\mathbf x_0)$ knowing input data has been decomposed into additive terms, supporting the identification and the interpretation of the contribution of different variability sources. Based on these formulas, several variance estimates independent of the noise distribution were provided for homoskedastic and heteroskedasic cases, for which a Python implementation was provided. Formulas and estimate-related theoretical results are supported by numerical simulations and empirical findings. 
Bias-reduced estimate $\hat \sigma^2_{BR}$ is likely uniformly better than $\hat \sigma^2_{NHo}$ in the homoskedastic case and should be prefer. In the heteroskedastic case, $\hat \sigma^2_{S2}$ and $\hat \sigma^2_{S3}$ are empirically shown to be better than other proposed estimates. Although these estimates are close to each other, $\hat \sigma^2_{S2}$ is computationally more efficient than $\hat \sigma^2_{S3}$. 

The paper also showed the possibility of constructing accurate CI for $f(\mathbf x_0)$ and $\mathbb E[\hat f(\mathbf x_0)]$
despite the non-parametric, non-linear, and random nature of ELM.
It provided a detailed explanation of the bias/variance contribution in CI estimation and highlighted that bias must be carefully consider to achieve satisfactory performances, especially in the regularized case which introduce significant bias. 
In particular, bias was traded against variance which can be estimated while preserving the predicting performance of the modelling, leading to credible uncertainty estimation. 
Also, as the variance estimates are distribution-free, it is reasonable to think that CI could be built with non-Gaussian noise distributional assumptions.

Several aspect of ELM uncertainty quantification still need to be investigated.
From a theoretical perspective, (asymptotical) normality of ELM (ensembles) should be proved or disproved. More generally, 
having an analytical expression for the distribution of $H$ could be very useful to develop estimation based on a single ELM. Additionally, random matrix theory --- which already provided theoretical results for ELM \citep{louart2018random}  --- should be investigated in the uncertainty quantification context.

Practically, prediction variance estimation is straightforward by adding $\hat\sigma_\varepsilon^2$ to the variance estimate in the homoskedastic case, while the noise variance could be estimated in the heteroskedastic case, e.g. with a second model \citep{Wan2014, Akusok2019}.
Prediction interval can also be constructed, assuming convenient noise distribution.
Future studies could also involve dependant data, e.g. by
adapting heteroskedasticity and autocorrelation consistent (HAC) estimations of the full noise covariance matrix, in the temporal or spatial cases \cite{davidson2004econometric, newey1986simple, kelejian2007hac}.







\section{Appendix}

\begin{proof}[Proof of equation \eqref{varELME}]
Recall that $W_1, \dots W_M$ are i.i.d..
Reusing Eq. (\ref{equ:linbias}), one have
\begin{align}\label{linbias2}
\begin{split}
\ECond{\hat  f (\mathbf x_0)}{W_1,  \dots,  W_M} 
&=
\frac{1}{M}\sum_{m=1}^M \ECond{\hat  f_m (\mathbf x_0)} {W_m} \\
&=\frac{1}{M}\sum_{m=1}^M \mathbf h_m ^T H_m^{\alpha} \mathbf f. \\
\end{split}
\end{align}
The law of total expectation yields 
\begin{align*}
\E{ \hat f (\mathbf x_0) - f (\mathbf x_0) }
&= \E{ \ECond{\hat  f (\mathbf x_0)}{  W_1,  \dots,  W_M } } - f (\mathbf x_0) \\
&= \frac{1}{M}\sum_{m=1}^M \underbrace{ \E{ \mathbf h_m ^T H_m^{\alpha} \mathbf f }}_{= \text{ cst}}- f (\mathbf x_0) \\
&=  \E { \mathbf h ^T H^{\alpha} \mathbf f} - f (\mathbf x_0) , 
    \end{align*}
and the bias still unchanged. 
Computing the first term of the law of total variance, one get
\begin{align*}
\VarCond{\hat f (\mathbf x_0) }{ W_1,  \dots,  W_M}
&= \VarCond{ \frac{1}{M}\sum_{m=1}^M \mathbf h_{m} ^T  H_m^{\alpha}  \mathbf y  }{ W_1,  \dots,  W_M }\\
&=  \frac{1}{M^2} \sum_{m,l = 1}^M \CovCond{ \mathbf h_m ^T  H_m^{\alpha}  \mathbf y}{ \mathbf h_l ^T  H_l^{\alpha}  \mathbf y }{ W_m,   W_l} \\
&=  \frac{1}{M^2} \sum_{m,l = 1}^M \mathbf h_m ^T H_m^{\alpha} \Var{\mathbf y}  H_l^{\alpha T}  \mathbf h_l  \\
&=  \frac{1}{M^2} \sum_{m,l = 1}^M \mathbf h_m ^T H_m^{\alpha} \Sigma H_l^{\alpha T} \mathbf h_l,
\end{align*}
which implies
\begin{align}\label{var1}
\begin{split}
\E{\VarCond{\hat f (\mathbf x_0) }{ W_1, \dots,  W_M}}
&=  \frac{1}{M^2} \sum_{m = 1}^M \underbrace{\E{ \mathbf h_m^T H_m^{\alpha} \Sigma H_m^{\alpha T} \mathbf h_m  }}_{= \text{ cst}} \\
& \quad +  \frac{1}{M^2}  \sum_{m \neq l } \E{ \mathbf h_m^T H_m^{\alpha}} \Sigma \E{ H_l^{\alpha T} \mathbf h_l } \\
&=  \frac{1}{M}  \E{ \mathbf h^T H^{\alpha} \Sigma H^{\alpha T} \mathbf h}\\
&\quad +  \frac{M-1}{M}  \E{ \mathbf h^T H^{\alpha}} \Sigma \E{H^{\alpha T} \mathbf h },
\end{split}
\end{align}
where one used the fact that the input weights and biases are drawn independently. 
Using (\ref{linbias2}), the second term of the law of total variance is
\begin{align}\label{var2}
\begin{split}
\Var{ \ECond{\hat f (\mathbf x_0) }{ W_1, \dots,  W_M} }
&=\frac{1}{M^2}\sum_{m,l=1}^M \Cov{ \mathbf h_m^T H_m^{\alpha} \mathbf f}{ \mathbf h_l ^T H_l^{\alpha} \mathbf f }\\
&=\frac{1}{M^2}\sum_{m=1}^M \underbrace{\Var{ \mathbf h_m^T H_m^{\alpha} \mathbf f}}_{=\text{ cst}}\\
&=\frac{1}{M} \Var{ \mathbf h^T H^{\alpha} \mathbf f },
\end{split}
\end{align}
as  $ \Cov{ \mathbf h_m^T H_m^{\alpha} \mathbf f}{ \mathbf h_l^T H_l^{\alpha} \mathbf f }$ vanishes when $m\neq l$, thanks again to the i.i.d. assumption on the weights. By summing Eq. \eqref{var1} and Eq. \eqref{var2}, the result is obtained. 
\end{proof}

\begin{proof}[Proof of equation \eqref{eq:CovBound}]
First, note that for all $m, l, k = 1, \dots M$,
\begin{align}\label{An:lemme1}
\begin{split}
\Cov{\mathbf{r}_m^T\mathbf{r}_m}{
\mathbf z^T_k
\mathbf z_l }
&=
\Cov{\ECond{\mathbf{r}_m^T\mathbf{r}_m}{W_m}}{
\ECond{\mathbf z^T_k
\mathbf z_l}{W_l, W_k} }\\
&\quad +
\E{\CovCond{\mathbf{r}_m^T\mathbf{r}_m}{
\mathbf z^T_k
\mathbf z_l }{W_m, W_l, W_k}}
\\
&= 
\Cov{ \mathbf{f}^T \mathbf{b}_{w,m}}{
\mathbf z^T_k
\mathbf z_l},\\
\end{split}
\end{align}
where for the first equality uses the law of total covariance, and the second equality uses equation \eqref{eq:ERSSW_homo}, the fact that expectation as no effect on a constant and that covariance between a random variable and a constant is null.
Also, using the covariance definition and the independence of weights between models, one have for all $k \neq l$,
\begin{align}\label{An:lemme2}
\begin{split}
\Cov{ \mathbf{f}^T \mathbf{b}_{w,k}}{
\mathbf z^T_k
\mathbf z_l}
&=
\E{ \mathbf{f}^T \mathbf{b}_{w,k}
\mathbf z^T_k\mathbf z_l}
-\E{ \mathbf{f}^T \mathbf{b}_{w,k}}
\E{\mathbf z^T_k\mathbf z_l}\\
&=
\E{ \mathbf{f}^T \mathbf{b}_{w,k}
\mathbf z^T_k} \boldsymbol{\mu}
-\E{ \mathbf{f}^T \mathbf{b}_{w,k}}
\E{\mathbf z^T_k}\boldsymbol{\mu}\\
&= \Cov{ \mathbf{f}^T \mathbf{b}_{w,k}}{
\mathbf z^T_k
\boldsymbol{\mu}},
\end{split}
\end{align}
Then, using equations \eqref{eq:homo_diagoff}, \eqref{eq:EstimNoise}, \eqref{An:lemme1} and \eqref{An:lemme2}, one obtains
\begin{align*}
\Cov{
\EstimNoise }{\widehat{\boldsymbol{\mu}}^T  \widehat{\boldsymbol{\mu}} -\frac{1}{M}\TrSmall{\hat Q} }
&= \frac{1}{M^2(M-1)(n-N)}\sum_{m=1}^M
\sum_{k \neq l} 
\Cov{\mathbf{r}_m^T\mathbf{r}_m}{
  \mathbf z^T_k
    \mathbf z_l }\\
&= \frac{1}{M^2(M-1)(n-N)}\sum_{m=1}^M
\sum_{k \neq l} 
\Cov{ \mathbf{f}^T \mathbf{b}_{w,m}}{
\mathbf z^T_k
\mathbf z_l}\\
&= \frac{2}{M^2(M-1)(n-N)}
\sum_{k \neq l} 
\Cov{ \mathbf{f}^T \mathbf{b}_{w,k}}{
\mathbf z^T_k
\mathbf z_l}
\\
&= \frac{2}{M^2(M-1)(n-N)}
\sum_{k \neq l} 
\Cov{ \mathbf{f}^T \mathbf{b}_{w,k}}{
\mathbf z^T_k
\boldsymbol{\mu}}
\\
&= \frac{2}{M(n-N)}
\Cov{ \mathbf{f}^T \mathbf{b}_{w}}{
\mathbf z^T
\boldsymbol{\mu}},
\end{align*}
where covariances vanish for $m\neq k,l$ in the third equality.
By using the Cauchy-Schwarz inequality on the last equation, one gets
\begin{equation*}
    \left| \;
    \Cov{
    \EstimNoise }{\widehat{\boldsymbol{\mu}}^T  \widehat{\boldsymbol{\mu}} -\frac{1}{M}\TrSmall{\hat Q} }\; \right|     
    \leq
    \frac{2}{M(n-N)}
    \sqrt{\Var{\mathbf{f}^T \mathbf{b}_{w}} \;
    \Var{\mathbf z^T
\boldsymbol{\mu}}}
\end{equation*}
from which equation \eqref{eq:CovBound} is obtained.

\end{proof}

\section*{Software Availability}
UncELMe --- Uncertainty quantification of Extreme Learning Machine ensemble --- is a Python package proposed on PyPI and GitHub (https://github.com/fguignard/UncELMe). It allows interested users to compute all variance estimates for Extreme Learning Machine ensemble discussed in the present paper. It is built within the scikit learn estimator framework, which enable the use of all convenient functionalities of scikit-learn \cite{pedregosa2011scikit}.
Noise estimation are also returned to enable building of prediction intervals.

\section*{Author Contributions}
F.G. 
conceived the main conceptual ideas, 
conduct investigations, 
developed the theoretical formalism and the methodology, 
performed the calculations, 
interpreted the computational results, 
wrote the original draft, 
and developed the Python software. 
M.K. 
carried out the supervision, 
project administration 
and funding acquisition. 
F.G., F.A. and M.K. 
discussed the results, 
provided critical feedback, 
commented, 
reviewed and edited the original manuscript, 
corrected the final version of the paper, 
and gave final approval for publication.

The original idea of this article stems in a conference paper \citep{guignard2020model} presented at the 28th European Symposium on Artificial Neural Networks, Computational Intelligence and Machine Learning (ESANN 2020) in Bruges, Belgium, from 2 to 4 October 2020. More precisely, in \citep{guignard2020model} equations \eqref{equ:modvar1} and \eqref{varELME} and naive estimates were presented for the non-regularized case only, with very few justifications and without any details. All other work presented in this paper is the result of an original research.

\section*{Acknowledgments}
This work is supported by the National Research Programme 75 “Big Data” (PNR75) of the Swiss National Science Foundation (SNSF), project no. 167285.
The authors are grateful to Prof. Dr. David Ginsbourger from Idiap Research Institute and University of Bern, Switzerland, for providing relevant insights on probability calculus. They also thank Dr. Mohamed Laib, Dr. Sylvain Robert and Dr. Jean Golay for the profitable discussions.

\section*{Declaration of interests}
The authors declare that they have no known competing financial interests or personal relationships that could have appeared to influence the work reported in this paper.

\bibliography{mybibfile}

\end{document}